# A Review of Autonomous Road Vehicle Integrated Approaches to an Emergency Obstacle Avoidance Maneuver

Evan Lowe          Levent Guvenç

(*lowe.500@osu.edu*)          (*guvenc.1@osu.edu*)

The Ohio State University, Automated Driving Lab

May 2021

## Abstract

As passenger vehicle technologies have advanced, so have their capabilities to avoid obstacles, especially with developments in tires, suspensions, steering, as well as safety technologies like ABS, ESC, and more recently, ADAS systems. However, environments around passenger vehicles have also become more complex, and dangerous. There have previously been studies that outline driver tendencies and performance capabilities when attempting to avoid obstacles while driving passenger vehicles. Now that autonomous vehicles are being developed with obstacle avoidance capabilities, it is important to target performance that meets or exceeds that of human drivers. This manuscript highlights systems that are crucial for an emergency obstacle avoidance maneuver (EOAM) and identifies the state-of-the-art for each of the related systems, while considering the nuances of traveling at highway speeds. Some of the primary EOAM-related systems/areas that are discussed in this review are: general path planning methods, system hierarchies, decision-making, trajectory generation, and trajectory-tracking control methods. After concluding remarks, suggestions for future work which could lead to an ideal EOAM development, are discussed.

**Keywords**: vehicle dynamics, obstacle avoidance, highway speeds, real-time, low surface mu, kinodynamic, decision-making, trajectory, tracking, machine learning, reinforcement learning, Markov decision process, neural networks, human driver

## Introduction

### Motivation

Though automotive safety technology has improved substantially over recent decades, there are still several accidents each year, and from those accidents many deaths result. Worldwide, approximately 1.25 million people are killed on roadways with more than half of the 3,700 fatalities being due to traffic crashes involving pedestrians, cyclists, and motorcyclists[1] (World Health Organization, 2018) (Centers for Disease Control and Prevention (CDC), 2020).

When taking a closer look at what the causes are for the reported accidents in the US, the National Motor Vehicle Crash Causation Survey (NMVCCS[2]), conducted from 2005 to 2007, showed that 94% of all vehicle crashes in the US were due to driver error, and the following table shows the breakdown of those driver-related errors (Singh S. , 2015)(Figure 1). Of those driver-related errors, the largest percentage (41%) of the human error was due to Recognition Errors, which can include driver inattention, internal and external distractions, and inadequate surveillance. This was followed by Decision Errors (33%) such as driving too fast for the existing road conditions, and misjudgment of either the gap between vehicles or other vehicles' speed as shown in Figure 1.

Amongst the Driver-Related Critical Reasons, all of them could be improved with the help of autonomous vehicles, depending on sensing, recognition, and performance capabilities of the applied systems. Additionally, when looking at the causes that are more directly related to environmental conditions, many of these would also be difficulties that autonomous vehicles would have to deal with.

---

[1] Vulnerable road users or VRUs are classified as pedestrians, cyclists, and motorcyclists by the World Health Organization.
[2] The NMVCSS is created by the National Highway Transportation Safety Administration (NHTSA) which is a governing body for United States federal safety regulations regarding road-going vehicles, and roadway standards.

The largest representative of the 52,000 Environment-Related causes of accidents were slick roads (50%), and another key cause is due to highway-related conditions (9%) (Singh S. , 2015). These Environment-Related causes are prominent in future work for emergency obstacle avoidance for autonomous vehicles. In other words, to help reduce crashes caused by slick roads and those at higher speeds (highway-related conditions), autonomous vehicle technology that utilizes emergency obstacle avoidance methodologies could help.

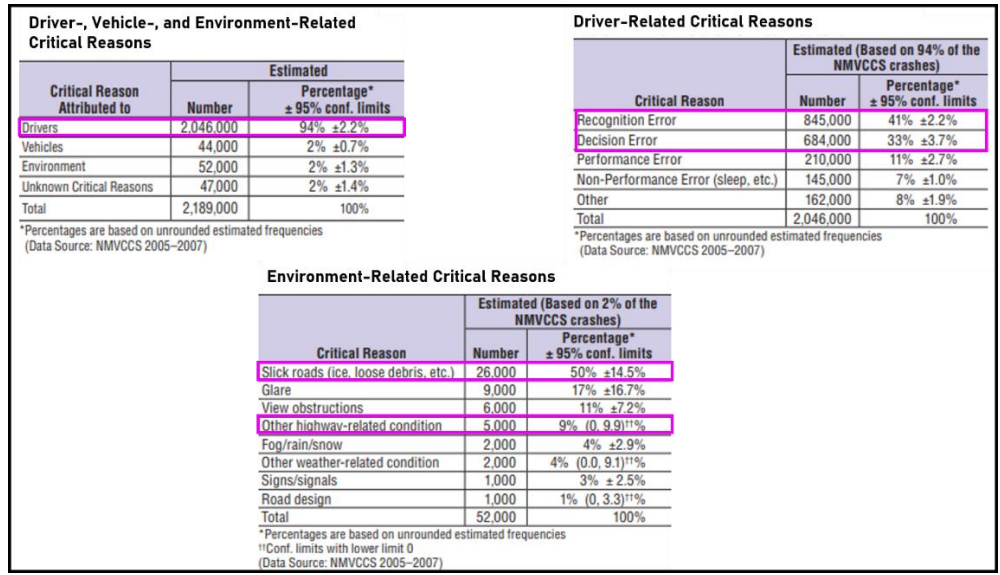

*Figure 1: NHTSA statistics regarding driver-caused crashes due to recognition errors, slick roads, and high speeds (Singh S. , 2015)*

These sobering statistics demonstrate that human driving error causes many car accidents, which are often fatal, even with the latest vehicle safety improvements. These statistics grow in severity as: a) vehicle speeds increase, and b) road and/or environmental conditions degrade.

One predominant takeaway from these collective statistics is that an Autonomous Road Vehicle (ARV) could be well-suited to safely deal with these driving situations that require evasive maneuvering, with high precision and accuracy, full utilization of the associated vehicle dynamics, and comprehensive real-time data regarding the outside environment. In this context, and ARV with a Level 3-5 autonomy, as defined by the Society of Automotive Engineers J3016 standard (Society of Automotive Engineers, 2016). An ARV that has an exclusive subsystem designed to handle emergency obstacle avoidance maneuvers (EOAMs) would be the ideal candidate to specialize in more severe ARV obstacle avoidance tasks.

## Unique Contributions

Currently, the literature is rich with reviews that approach ARV path planning in addition to low and high-speed obstacle avoidance. From a broad sense, autonomous path planning and obstacle avoidance are covered in autonomous robotics ( (Kamil, Tang, Khaksar, Zulkifli, & Ahmad, 2015), (Elbanhawi & Simic, 2014), (Laumond, Sekhavat, &

Lamiraux, 1998) ). Alternatively, ARVs are inherently non-holonomic[3] (Whittaker, 1904), (Laumond, Sekhavat, & Lamiraux, 1998) and in modern path planning, kinodynamic[4] (Canny, Donald, Reif, & Xavier, 1988) as well.

These past reviews cover the motion planning for both non-holonomic and kinodynamic vehicles, such as automotive passenger cars ( (Katrakazas, Quddus, Chen, & Deka, 2015), (González, Pérez, Milanés, & Nashashibi, 2015), (Paden, Čáp, Yong, Yershov, & Frazzoli, 2016), and (Claussmann, Revilloud, Gruyer, & Glaser, 2019) ). These reviews are comprehensive in detailing ARV systems (Gordon & Lidberg, 2015), and particularly focused on path planning in an urban sense. Claussmann extends these reviews into general path planning at highway autonomous driving, that does not include emergency maneuvers (Claussmann, Revilloud, Gruyer, & Glaser, 2019). However, a review that explores the considerations of emergency obstacle avoidance at speeds that exceed those of an urban setting[5], does not currently exist in the literature.

The purpose of this review is to broach the necessary considerations for an EOAM at speeds that exceed those in an urban setting. These considerations include not only the specifics of the EOAM subsystem itself, but also the related subsystems that directly and indirectly interact with the EOAM, as well as the ARV as an entire system. An ideal EOAM at highway speeds is that which is feasible to be executed by the ego vehicle, safe (for the occupants, surrounding vehicles, and pedestrians), real-time capable, robust to changes in the environment and vehicle states, comfortable for the vehicle occupants, and well-integrated into the rest of the entire ARV control architecture and physical systems.

Areas of obstacle avoidance that involve prediction and artificial intelligence-related strategies, such as machine learning, reinforcement learning, neural networks, game theory, the Markov decision process (Schwarting, Alonso-Mora, & Rus, 2018), or other general artificial intelligence methods will not be discussed in this review. While these areas are critical for advancement of ARV technology, these methods are out of the scope of this review.

### Review Layout
This review highlights the relevant ARV systems which can work together to achieve an ideal EOAM. The organizing structure of this review is as follows: Background in General ARV Systems and Vehicle Safety Technologies, General ARV Path Planning and System Hierarchies, Actuator and Vehicle Modeling, Decision-Making Processes for ARVs, Trajectory Generation for Emergency Maneuvers, and Trajectory-Tracking Control Strategies. This review ends with a Conclusion, and recommendations for Future Work.

# Background
### General ARVs
The context of this manuscript centers around what can be defined as an emergency obstacle avoidance maneuver (EOAM). An EOAM is an evasive action that a driver or driving system must take while operating an automotive vehicle, to safely avoid an in-road obstacle within the constraints of the outside environment and local traffic rules. When considering an effective EOAM for ARVs, it is important to consider the entire ARV as a functional system. This macro-system view is critical, due to the inherent time-dependence within any obstacle avoidance strategy, as one of the factors that controls success or failure.

There are many ways in which an ARV broad system hierarchy can be viewed[6], but often the key components are demonstrated well in the diagram below in Figure 2:

---

[3] Non-holonomic systems are those which include non-integrable constraints, and thus the number of independent configuration variables exceed the available degrees of freedom (DOFs).

[4] Kinodynamic planning represents the set of motion problems which are subject to simultaneous kinematic (and dynamic constraints. Kinematic constraints may be avoiding obstacles, while dynamic constraints could be limitations on such things as velocity, acceleration, and force.

[5] As defined by Claussmann, speeds included in highway driving are those which meet or exceed 60 km/h.

[6] Samples of the various hierarchies will be discussed in the ARV System Hierarchies section, below.

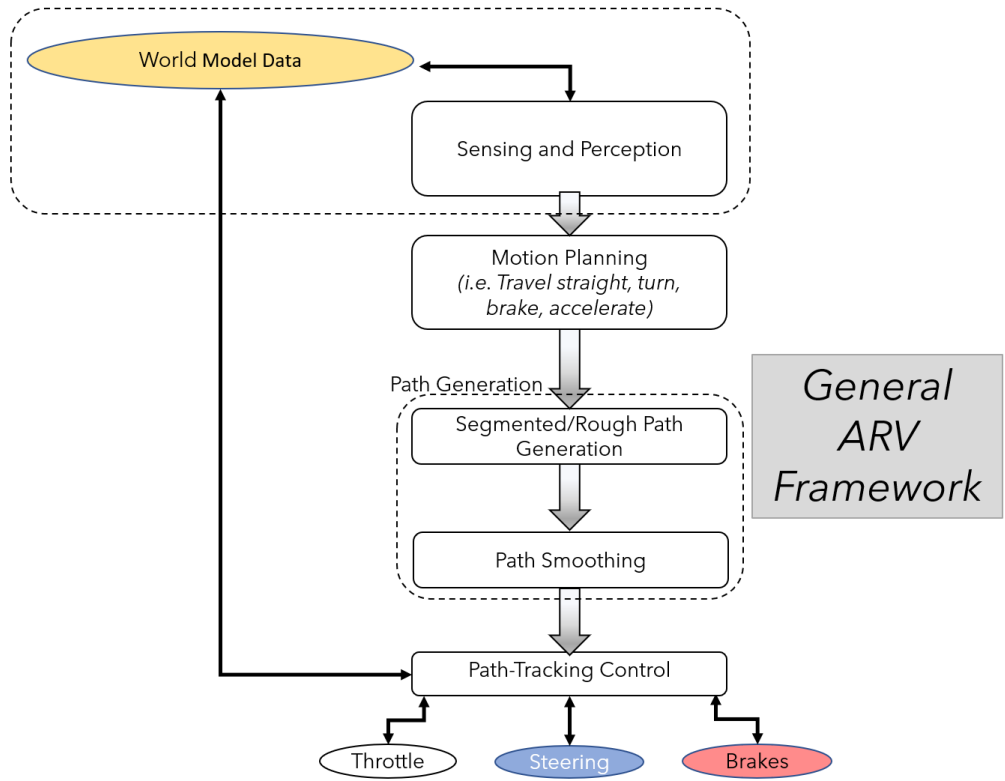

*Figure 2: Broad system level hierarchy and description of an ARV*

The key ARV system components are listed here as:

- World Model
- Sensing and Perception
- General Planning and Decision-Making
- Rough Path Generation and Smoothing
- Path-Tracking Control
- Actuation (steering, throttle, brakes)

The World Model is described as a system which merges incoming sensor and communication data to provide an accurate and up-to-date representation of the ARV's environment, at any time (Furda & Vlacic, 2010), (Furda & Vlacic, 2011). Each of these listed key ARV system components will be described in detail, in the ARV System Hierarchies section, below.

## General Obstacle Avoidance versus EOAMs

A general autonomous obstacle avoidance maneuver can be categorized as one which includes minor path updates and does not require exceptional dynamic performance (Ziegler et al., 2014), (Wang, Tota, Aksun-Guvenc, & Guvenc, 2018). Specifically, an emergency obstacle avoidance maneuver (EOAM) at highway speeds[7] distinguishes itself from general obstacle avoidance, in that the ego vehicle is traveling at higher speeds or lower tire/road surface mu[8], than those typically used in an urban setting and dry conditions, requiring more rapid decision-making and actuator inputs.

---

[7] Highway speeds are defined by those used in the Euro NCAP Highway Assist Systems Test & Assessment Protocol, to be 50–130 kph h (31.1-80.8 mph) ( (European New Car Assessment Programme, 2020)

[8] Surface mu values for roads similar to asphalt are widely considered as follows: 0.8-1 dry surface, 0.7-0.8 for wet surfaces, 0.2-0.3 for pack snow, 0.2-0.5 for loose snow/slush, and ice can range from 0.05-0.3 depending on conditions. Low surface mu in the context of this review can range from wet to icy conditions (Wallman & Åström, 2001).

These needs for more rapid decision-making and actuator inputs also require exceptional vehicle dynamic performance and robust controller processing power to achieve a safe result in real-time. Indirectly, the EOAM also requires rapid Sensing and Perception system capabilities so that accurate and high frequency World Model data may be input to the Decision-Making and Trajectory-Tracking Control systems.

## History of Safety Technologies

### Early Vehicle Safety Technologies

Before ARVs were in development to improve driving safety (among other goals), vehicle control systems designed to reduce the severity of accidents, or avoid them completely, were introduced. These passive safety control systems include anti-lock braking systems (ABS), traction control systems (TCS), and what is commonly called electronic stability control (ESC). These listed passive safety applications require a multitude of sensors which can work independently, but often provide coordinated information to electronic controllers, which are dedicated to managing the vehicle's steering, throttle, and brakes (Bengler, et al., 2014). ABS is used when the driver is applying the brake, to maintain some steering control and avoid locking the brakes during purely longitudinal stops (Leiber & Czinczel, 1984), (Dincmen, Aksun-Guvenc, & Acarman, 2014). TCS is active when the driver is applying the throttle and torque must be reduced at the driving wheels to maintain optimal traction/reduce excessive wheel slip (Maisch, Jonner, & Sigl, 1987). The purpose of ESC is to maintain yaw stability during any turn maneuvers, and be combined with TCS, or ABS depending on the maneuver (van Zanten, Erhardt, & Pfaff, 1995), (Emirler, et al., 2015), (Aksun-Guvenc, Guvenc, & Karaman, Robust MIMO Disturbance Observer Analysis and Design with Application to Active Car Steering, 2010), (Aksun-Guvenc, Bunte, Odenthal, & Guvenc, 2004), (Aksun Guvenc & Guvenc, 2002), (Aksun-Guvenc, B; Guvenc, L; Karaman, S, 2009).

### Current Vehicle Safety Technologies

While the passive safety systems described above can be significant in reducing the severity of vehicle collisions, and possibly aid drivers in avoiding collisions all-together, the human driver is ultimately responsible for control of the vehicle during emergency events, even with these safety systems in place. The major next step in vehicle safety technology[9] was the development of automatic driver assistance systems, or ADAS. ADAS can provide a host of vehicle control options including lane-keeping assist (LKAS) (Cantas & Guvenc, 2018), (Coskun, Tuncer, Karsligil, & Guvenc, 2010), automatic cruise control (ACC) (Kural & Aksun-Guvenc, 2010), road departure medication (RDM), and traffic sign/signal and signal detection (TSD), but regarding emergency maneuvers, the primary ADAS feature has been automatic emergency braking or AEB ( (Ziebinski, Cupek, Grzechca, & Chruszczyk, 2017), (Bengler, et al., 2014), (You, et al., 2015), (Fors, Olofsson, & Nielsen, 2018)) (Gordon & Lidberg, 2015).

AEB and other ADAS systems utilize vehicle sensing through devices like radars, lidars, laser scanners, and cameras to recognize dangerous obstacle in the road ahead of the ego vehicle (Haus, Sherony, & Gabler, 2019) (Hamberg, Hendriks, & Bijlsma, 2015), with one potential sensor layout shown by (Gong, 2016)[10] in Figure 3.

---

[9] These can be all be classified as active safety systems because they are often better in sensing and reacting to road activity than human drivers (Bengler, et al., 2014). However, LKAS, and ACC can also be defined as convenience features that aid in lowering the driver's mental fatigue while driving.

[10] Here, the six regions-of-interest are denoted as front left (FL), front ego (FE), front right (FR), rear ego (RE), and rear right (RR)

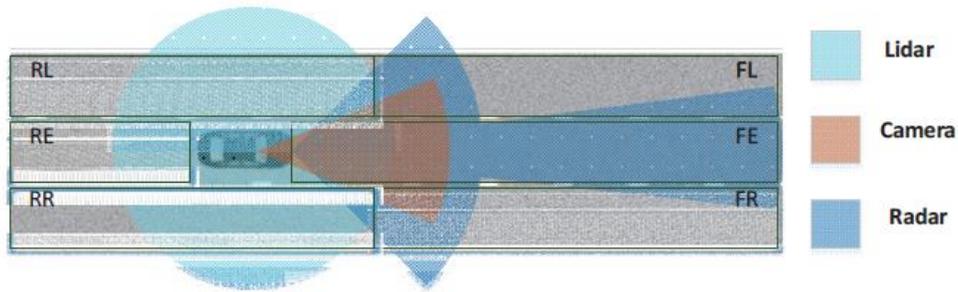

If the ego vehicle is approaching the dangerous obstacle and the driver does not initiate braking, with proper sensing and recognition, the AEB system will activate by applying the brakes until the vehicle comes to a complete stop, or slows down to a speed which AEB is no longer deemed necessary by the AEB controller (Hamberg, Hendriks, & Bijlsma, 2015).

## Future Vehicle Safety Technologies

While ADAS systems can independently provide autonomous control of steering, throttle or brakes under certain specific conditions, they are still classified as Level 1 or 2 systems (Guvenc, Aksun-Guvenc, & Emirler, Chapter 35 Connected and Autonomous Vehicles, 2016) (Society of Automotive Engineers, 2016) (Figure 4). These are systems that support the driver who must be constantly aware of the vehicle behavior and must supervise the ADAS outputs, in the case human control must be resumed. Truly ARV systems can be classified demonstrating SAE Automation Levels 3- 5: the vehicle is driving itself whether there is a driver in the driver's seat or not. Levels 3 and 4 driving has limited conditions by which the autonomous system can maintain control of the vehicle, whereas level 5 can operate under all road conditions and does not ever need driver intervention.

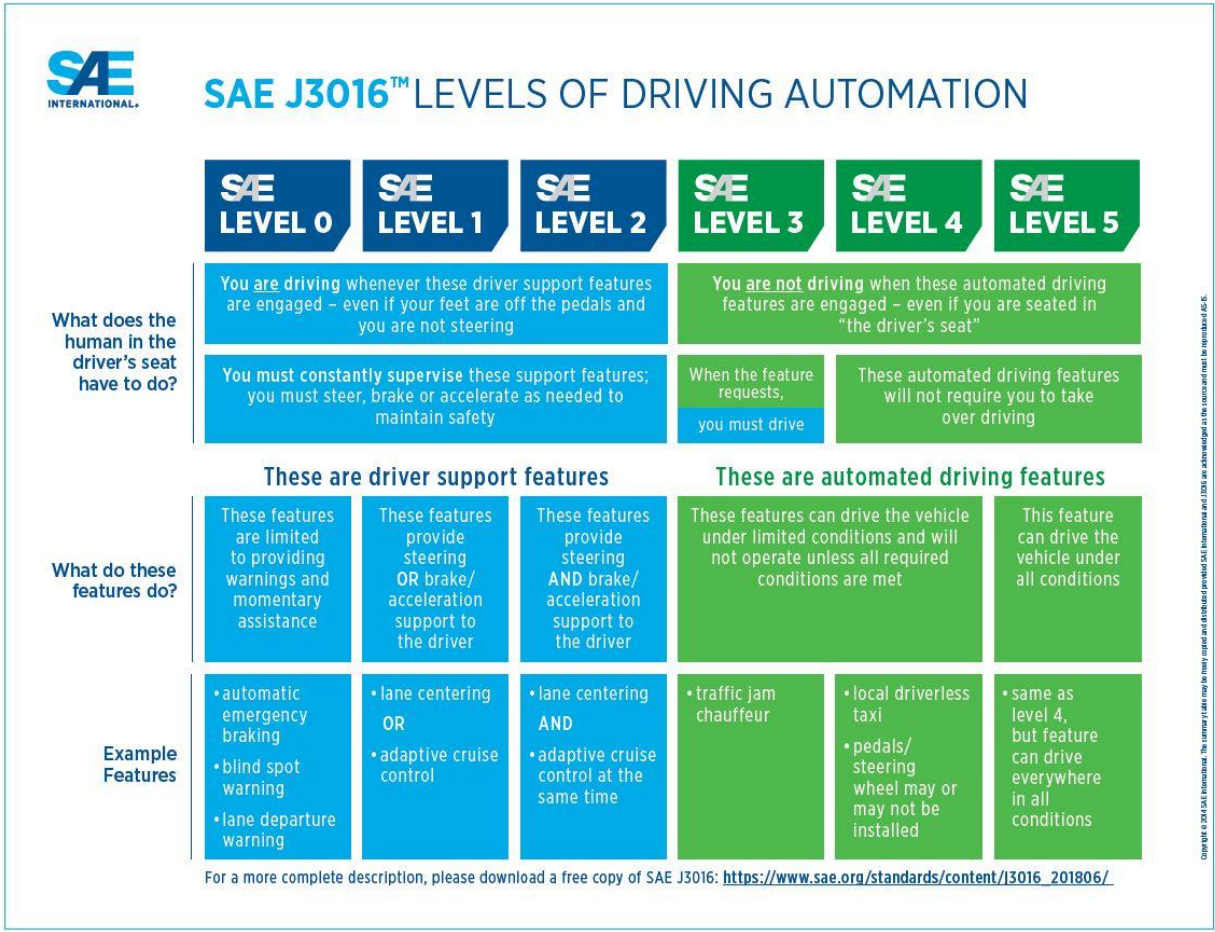

*Figure 4: SAE J3016 Levels of Driving Automation Infographic (Society of Automotive Engineers, 2016)*

This brief recap of early, current and future automotive safety technologies serves as a basis to understand the ARV systems relating to vehicle safety, and especially those involved with EOAMs. The rest of this review is dedicated to diving into the subsystem details of ARV EOAM applications while considering the entire ARV at the system level. The goal is to unify EOAM-related subsystems as discussed in the literature, to emphasize the need for all of these subsystems to attain successful outputs and communication at the system level, when attempting to execute an EOAM.

# Emergency Obstacle Avoidance Strategies and Systems

While this overall review of Emergency Obstacle Avoidance Maneuvers (EOAMs) covers the relevant state-of-the-art for EOAM strategies, it extends beyond the literature for the EOAMs themselves. Rather, other subsystems that are directly and indirectly related to EOAMs are also covered because they are believed to be impactful on the system design of EOAMs. Additionally, it is important that the EOAM is not considered as an isolated entity, because its success depends on how it interacts with other subsystems within an ARV system, in its entirety. Therefore, the Review of Emergency Obstacle Avoidance Strategies is broken up into the following relevant subsections: General ARV Path Planning, Decision-Making Processes for ARVs, Trajectory Generation for Emergency Maneuvers, Trajectory-Tracking Control Strategies, and Kinodynamic Considerations. The initial discussion will begin with General ARV Path Planning, which provides a macro-scope for path planning of ARVs.

## General ARV Path Planning

At a high level, the primary methods for general path planning of an Autonomous Road Vehicle (ARV) involve guidance of an ARV through a series of waypoints, to reach its final destination in a feasible manner (Katrakazas,

Quddus, Chen, & Deka, 2015), (Guvenc, Aksun-Guvenc, Zhu, & Gelbal, 2021). While the general autonomous path planning is reviewed in this manuscript briefly, there is an important difference between general path generation and trajectory generation. General ARV motion directives generate paths rather than trajectories, and path generation focuses on general spatial location and motion while trajectory generation critically relies on time-related velocity and acceleration (Paden, Čáp, Yong, Yershov, & Frazzoli, 2016), (Peng, et al., 2020), (Li, Zhu, Aksun-Guvenc, & Guvenc, 2021).

General ARV path planning is fundamentally important, as maintaining a desired path occupies most of the motion directives of an ARV, albeit at generally lower speeds that do not require highly dynamic motions. This general path planning can occur on urban and non-urban roads, maintaining lane boundaries traffic rules, while avoiding obstacles, including other road users and road irregularities at speeds lower than those on typical highways ( (Katrakazas, Quddus, Chen, & Deka, 2015), (Ziegler et al., 2014)). Part of this real-time planning includes defining a search or configuration space, which (Katrakazas, Quddus, Chen, & Deka, 2015) define as the set of entire possible independent attributes that can describe the ego vehicle's unique position and orientation, according to a fixed coordinate system (Katrakazas, Quddus, Chen, & Deka, 2015), (Ding Y. , et al., 2020). A summary of these lower speed path planning methodologies to define a search/configuration space, in addition to emphasis on key terminology relating to the real-time nature that is involved in the planning[11], can be seen below

.

| Representation | Advantages | Disadvantages |
|---|---|---|
| Voronoi Diagrams | • Completeness<br>• Maximum distance from obstacles | • Limited to static environments<br>• Discontinuous edges |
| Occupancy Grids Cost Maps | • Fast discretisation<br>• Small computational power[a] | • Problems with vehicle dynamics<br>• Errors in the presence of obstacles |
| State lattices | • Efficiency without increasing computational time[b]<br>• Pre-computation of edges is possible | • Problems with curvature<br>• Restrict motion<br>• Difficulties in dealing with evasive manoeuvres |
| Driving corridors | • Continuous collision free space for the vehicle to move | • Computational cost[c]<br>• Constraints on motion |

[a] Computational power refers to computations needed to construct the cells and estimate their costs. The space, in which the planning problem is solved, is discretised. Furthermore, the number of attributes needed to define each of the cells is small (the attributes just need to show if the cell is occupied or not, plus the cost of traversing the cell). As a result, the dimensions of the state matrix of each of the cells are manageable in real-time.

[b] Similar to (a), computational time refers to computations needed to construct the lattice: Because of the predefined shape of the curve with which the lattice is constructed and the pre-computation of edges, the space for planning is discretised and thus less time is needed to find the correct solution.

[c] Computational cost for driving corridors is analysed in footnote 3.

*Figure 5: Comparison of methodologies to define a search space (Katrakazas, Quddus, Chen, & Deka, 2015)*

Extending beyond the configuration space, (González, Pérez, Milanés, & Nashashibi, 2015) organized and summarized some of the main groups of general ARV path/motion planning, such as Dijkstra's and those from the A* family, but also later techniques such as those from the RRT family, function optimization, and various geometric methods (line and circle, clothoids, polynomials, etc.), in Figure 6.

---

[11] The footnote 3 which Katrakazas et al., 2015 note is quoted as follows, "The continuous nature of driving corridors, leads to an exponential increase in the dimensions of state vector for each one of the coordinates included in the driving corridor. Thus, at each time moment a large number of attributes need to be calculated for each of the coordinates, necessitating more computational resources." (Katrakazas, Quddus, Chen, & Deka, 2015)

| Algorithm group | Technique | Technique description | Implemented in |
|---|---|---|---|
| Graph search based planners | Dijkstra's Algorithm | Known nodes/cells search space with associated weights | [27], [51], [52], [53] |
| | | Grid and node/cells weights computation according to the environment | [28], [26] |
| | A* algorithm family | Anytime D* with Voronoi cost functions | [10], [16], [30] |
| | | Hybrid-heuristics A* | [36], [54] |
| | | A* with Voronoi/Lattice environment representation | [37], [35], [51] |
| | | PAO* as in [55] | [56] |
| | State Lattices | Environment decomposed in a local variable grid, depending on the complexity of the maneuver. | [16], [30], [46], [57] |
| | | Spatio-temporal lattices (considering time and velocity dimensions) | [50], [58], [39], [59], [47], [60] |
| Sampling based planners | RRT | Physical and logical bias are used to generate the random-tree | [61], [62], [63] |
| | | Anytime planning with RRT* | [64], [40], [53], [65]. |
| | | Trajectory coordination with RRT | [66] |
| Interpolating curve planners | Line and circle | Road fitting and interpolation of known waypoints | [41], [67] |
| | Clothoid Curves | Piecewise trajectory generation with straight, clothoid and circular segments | [68], [42], [69], [70],[71], [72] |
| | | Off-line generation of clothoid primitives from which the best will be taken in on-line evaluation | [73], [74], [14] |
| | Polynomial Curves | Cubic order polynomial curves | [75], [50] |
| | | Higher order polynomial curves | [76],[77], [78], [79], [80] |
| | Bézier Curves | Selection of the optimal control points location for the situation in hand | [81], [44], [82], [83], [79], [84] |
| | | Rational Bézier curves implementation | [85], [86] |
| | Spline Curves | Polynomial piecewise implementation | [15], [76], [28] |
| | | Basis splines (b-splines) | [87], [88], [89] |
| Numerical optimization approaches | Function optimization | Trajectory generation optimizing parameters such as speed, steering speed, rollover constraints, lateral accelerations, jerk (lateral comfort optimization), among others | [90], [91], [54], [5], [38], [92] |

*Figure 6: González summary of motion planning techniques (González, Pérez, Milanés, & Nashashibi, 2015)*

A list of advantages and disadvantages for the methods listed in Figure 6 were also provided in another table, as shown in Figure 7 below (González, Pérez, Milanés, & Nashashibi, 2015).

| Technique | Advantages | Disadvantages |
|---|---|---|
| Dijkstra's algorithm | Finds the shortest path in a series of nodes or grid. Suitable for global planning in structured and unstructured environments (Fig. 2a). | The algorithm is slow in vast areas due to the important amount of nodes. The search is not heuristic. The resulting path is not continuous. Not suitable for real time applications. |
| A* family | Based on the Dijkstra algorithm. The search is heuristic reducing computation time (Fig. 2d). | The resulting path is not continuous. The heuristic rule is not straightforward to find most of the times. |
| State lattices | Able to handle several dimensions (position, velocity, acceleration, time). Suitable for local planning and dynamic environments (Fig. 2c). | Computationally costly due to the evaluation of every possible solution in the database. The planner is only resolution complete (lattice discretization). |
| RRT family | Able to provide a fast solution in multi-dimensional systems. The algorithm is complete and always converges to a solution (if there is one and given enough time). Suitable for global and local planning, see Fig. 2e. | The resulting trajectory is not continuous and therefore jerky. The optimality of the path strongly depends on the time frame for the RRT* case. |
| Interpolating curve planner | Optimization of the curvature and smoothness of the path is achieved through the implementation of CAGD techniques (compared here below). Suitable for local planning oriented to comfort and safety in structured environments. | Depends on a global planning or global waypoints. Time consuming when managing obstacles in real time because the optimization of the path and consideration of road an ego-vehicle constraints. |
| Line and circle | Low computational cost. Simple to implement. Assures the shortest path for a car-like vehicle, see Fig. 2f. | The path is not continuous and therefore jerky, making non-comfortable transitions between segments of the path. The planner depends on global waypoints. |
| Clothoids | Transitions to and from curves are done with a linear change in curvature. Highways and road designs implement these curves. Suitable for local planning (see Fig. 2g). | Time consuming because of the integrals that define the curve. The curvature is continuous but not smooth (linear behavior). The planner depends on global waypoints. |
| Polynomials | Low computational cost. Continuous concatenations of curves are possible (Suitable for comfort). | Curves implemented are usually of 4th degree or higher, difficulting the computation of the coefficients to achieve a determined motion state. |
| Béziers | Low computational cost. Intuitive manipulation of the curve thanks to the control points that define it. Continuous concatenations of curves are possible (Suitable for comfort). See Fig. 2i. | Loss of malleability when increasing the curve degree, as well as the computation time increases (Thus, more control points have to be evaluated and correctly placed). The planner depends on global waypoints. |
| Splines | Low computational cost. The result is a general and continuous curvature path controlled by different knots (see Fig. 2j). | The solution might not be optimal (from the road fitness and curvature minimization point of view) because its result focuses more on achieving continuity between the parts than malleability to fit road constraints. |
| Function Optimization | Road and ego-vehicle constraints as well as other road users can be easily taken into account (Fig. 2b). | Time consuming since the optimization of the function takes place at each motion state. Therefore, the optimization is stopped at a given time horizon. The planner depends on global waypoints. |

*Figure 7: González comparison of advantages and disadvantages of various general motion planning techniques (González, Pérez, Milanés, & Nashashibi, 2015)*

To aid both Figure 6 and Figure 7, an illustrative view of the various general motion planning techniques from (González, Pérez, Milanés, & Nashashibi, 2015) which were reviewed can be seen in (Figure 8).

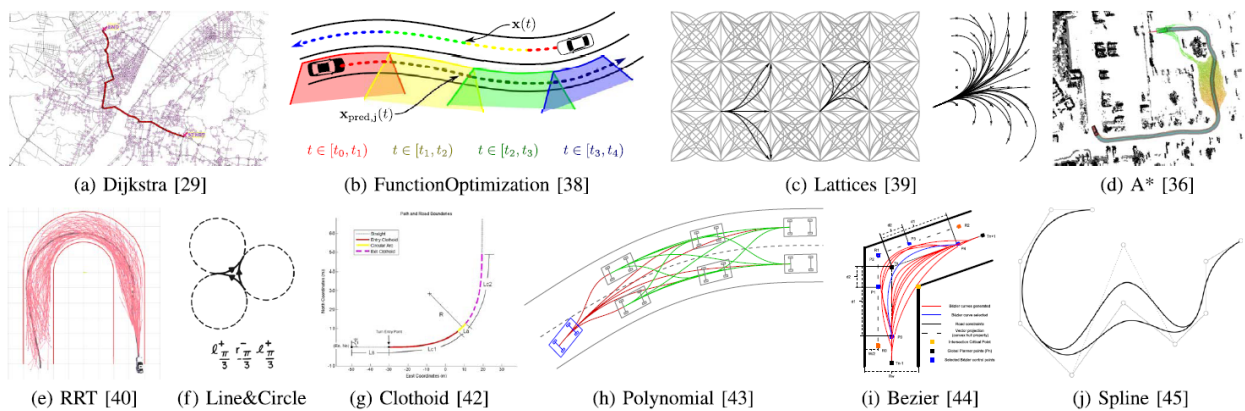

*Figure 8: General motion planning technique diagrams from González and referencing authors (González, Pérez, Milanés, & Nashashibi, 2015)*

As an explanation, in Figure 8, "Planning algorithms as presented in the literature. (a) Representation of a global path by the Dijkstra algorithm in [29]➔ (Li, Zeng, Yang, & Zhang, 2009). (b) Trajectory optimization from [38]➔ (Ziegler, Bender, Dang, & Stiller, 2014), taking into account a vehicle in the other lane. (c) Lattices and motion primitives as presented in [39]➔ (Ziegler & Stiller, 2009). (d) Hybrid A∗ as implemented in the DARPA Challenge by Junior [36]➔ (Montemerlo, et al., 2008). (e) RRT∗ as showed in [40]➔ (Jeon, et al., 2013). (f) Optimal path to turn the vehicle around as proven in [41]➔ (Reeds & Shepp, 1990). (g) Planning a turn for the Audi TTS from Stanford [42]➔ (Funke, et al., 2012). (h) Different motion states, planned with polynomial curves as presented in [43]➔ (Xu, Wei, Dolan, Zhao, & Zha, 2012). (i) Evaluation of several Bézier curve in a turn, as showed in [44]➔ (González, Pérez, Lattarulo, Milanés, & Nashashibi, 2014). (j) Spline behavior when a knot changes place, as presented in [45]➔ (Farouki, 2008)."

Also, as an important note, methods (f)-(j) as listed by (González, Pérez, Milanés, & Nashashibi, 2015) (Figure 8), are some of the common trajectory planning methods that can be used for EOAMs, in addition to general ARV path planning.

A different review by (Paden, Čáp, Yong, Yershov, & Frazzoli, 2016), addressed general path planning for ARVs, with an emphasis on model assumptions, computational completeness, potential level of optimality, time complexity, and anytime capable Figure 9.

| | Model assumptions | Completeness | Optimality | Time Complexity | Anytime |
|---|---|---|---|---|---|
| Geometric Methods | | | | | |
| Visibility graph [33] | 2-D polyg. conf. space, no diff. constraints | Yes | Yes [a] | $O(n^2)$ [68] [b] | No |
| Cyl. algebr. decomp. [76] | No diff. constraints | Yes | No | Exp. in dimension. [65] | No |
| Variational Methods | | | | | |
| Variational methods (Sec IV-C) | Lipschitz-continuous Jacobian | No | Locally optimal | $O(1/\epsilon)$ [77] [k,l] | Yes |
| Graph-search Methods | | | | | |
| Road lane graph + Dijkstra (Sec IV-D1) | Arbitrary | No [c] | No [d] | $O(n + m \log m)$ [78] [e,f] | No |
| Lattice/tree of motion prim. + Dijkstra (Sec IV-E) | Arbitrary | No [c] | No [d] | $O(n + m \log m)$ [78] [e,f] | No |
| PRM [79] [g] + Dijkstra | Exact steering procedure available | Probabilistically complete [80] [*] | Asymptotically optimal [*] [80] | $O(n^2)$ [80] [h,f,*] | No |
| PRM* [80], [81] + Dijkstra | Exact steering procedure available | Probabilistically complete [80]–[82] [*,†] | Asymptotically optimal [80]–[82] [*,†] | $O(n \log n)$ [80], [82] [h,f,*,†] | No |
| RRG [80] + Dijkstra | Exact steering procedure available | Probabilistically complete [80] [*] | Asymptotically optimal [80] [*] | $O(n \log n)$ [80] [h,f,*] | Yes |
| Incremental Search | | | | | |
| RRT [83] | Arbitrary | Probabilistically complete [83] [i,*] | Suboptimal [80] [*] | $O(n \log n)$ [80] [h,f,*] | Yes |
| RRT* [80] | Exact steering procedure available | Probabilistically complete [80], [82] [*,†] | Asymptotically optimal [80], [82] [*,†] | $O(n \log n)$ [80], [82] [h,f,*,†] | Yes |
| SST* [84] | Lipschitz-continuous dynamics | Probabilistically complete [84] [†] | Asymptotically optimal [84] [†] | N/A [j] | Yes |

*Figure 9: Paden et al., comparison table of general path planning methods (Paden, Čáp, Yong, Yershov, & Frazzoli, 2016)[12]*

It should be noted that while (Paden, Čáp, Yong, Yershov, & Frazzoli, 2016) introduce trajectory planning as the motion planning problems for dynamic environments, they do not include maneuvers which can be categorized as an EOAM. Still, the methodologies and considerations that they explain with regard to general ARV motion planning are comprehensive and mathematically rigorous.

In a form that was similar to (Paden, Čáp, Yong, Yershov, & Frazzoli, 2016), but with an emphasis on Sample-Based Planning (SBP) for general autonomous robot planning (not specific to automotive vehicle applications) motion, (Elbanhawi & Simic, 2014) provide an efficient summary of the SBP parameters and heuristics that they reviewed. However, within the scope of this manuscript and especially ARVs, the most relevant section of this table is that of the Local Planning methods, which directly relate to EOAMs (Figure 10).

**Local Planning**

| Method | Reference | Remark |
|---|---|---|
| Circle, arcs and lines | [154, 155] | Path is not continuous |
| Clothoids | [148, 156] | Difficult to compute online |
| Splines | [104, 149] | Curvature control is challenging |
| Polynomials | [143] | Difficult to compute online |

*Figure 10: Local Planning methods for autonomous robots, summary (Elbanhawi & Simic, 2014)*

Within Figure 10, the supporting authors are as follows 154: (Bui, Boissonnat, Soueres, & Laumond, 1994), 155: (Reeds & Shepp, 1990), 148: (Kanayama & Hartman, 1997), 156: (Fraichard & Scheuer, 2004), 104: (Yang K. , 2013), 149: (not published), 143: (Papadopoulos, Poulakakis, & Papadimitriou, 2002)

A review which also covered an extensive range of ARV path planning, though with a scope narrowed to basic highway driving was that by (Claussmann, Revilloud, Gruyer, & Glaser, 2019). In this review, a definition for highway driving was listed as having longitudinal speeds which are 60km/h or greater, on lane-divided roads[13] with shapes including straight lines, clothoids, and/or circles with small curvature, only motorized vehicles that adhere to the same driving rules, and whose motion is restricted to constant speed, acceleration, braking, and left or right lane changes (Claussmann, Revilloud, Gruyer, & Glaser, 2019). Additionally, the authors mention that the five key characteristics of ARV motion planning are (i) state estimation, (ii) time evolution, (iii) action planning, (iv) criteria optimization, and (v) compliance with constraints.

The review by (Claussmann, Revilloud, Gruyer, & Glaser, 2019) highlighted many autonomous driving details which are relevant to general ARV path planning and EOAMs. In the state-of-the art portion of their review, they offer first classifications for various algorithms, and then reflect details of the more popular and recent approaches and applications of those algorithms, with respect to autonomous highway driving. They, then, provide a helpful summary of the relative merits of each of the recent algorithm approaches and applications in a Comparison Table for Highway Applications of Motion Planning Methods, as seen in Figure 11 (Claussmann, Revilloud, Gruyer, & Glaser, 2019).

---

samples/algorithm iterations; *i*: for certain variants; *j*: not explicitly analyzed; *k*: $\epsilon$ is the required distance from the optimal cost; *l*: faster rates possible with additional assumptions; ∗: shown for systems without differential constraints; †: shown for some class of nonholonomic systems. (Paden, Čáp, Yong, Yershov, & Frazzoli, 2016)

[13] Claussmann defines lane-divided roads as those with unidirectional flow (opposing directions of travel being separated by a median strip)

*Figure 11: Claussmann Comparison Table for Highway Applications of Motion Planning Methods (Claussmann, Revilloud, Gruyer, & Glaser, 2019)*

Within the table from Figure 11, the legend for the relative ratings of the various methods are as follows: '—' very inappropriate, '−' inappropriate, '~' intermediary, '+' appropriate, '++' very appropriate. There is also a legend for the authors' designations of the use cases applicable for each motion planning method in their table. It should be noted that motion planning methods from (Claussmann, Revilloud, Gruyer, & Glaser, 2019) which are most relevant to EOAMs as noted in the current manuscript, are Parametric Curves Semi-Parametric Curves, Mathematical Optimization, and AI Logic.[14]

In summary, the reviews described above, offered value and substantive information regarding general ARV path/motion planning (Katrakazas, Quddus, Chen, & Deka, 2015), (Elbanhawi & Simic, 2014), (González, Pérez, Milanés, & Nashashibi, 2015), (Claussmann, Revilloud, Gruyer, & Glaser, 2019). The information offered in these reviews represent the majority of path planning that an ARV will likely conduct under normal circumstances, but do not fully apply to the circumstances of an EOAM. However, the authors still offer details of planning methodologies which are relevant when considering the formulation of an EOAM for an ARVs.

## ARV System Hierarchies

To understand the system-level controller logic that prompts the execution of an EOAM, it is important to understand the system hierarchies involved within the entire ARV system. This is due to the nature in which the ARV attains data regarding the surrounding environment, processes that data, and makes decisions, directly influences the execution,

---

[14] AI logic is primarily out of the scope of this current manuscript, however, some of the decision-making aspects of this methodology will be broached in later sections

timing, duration, and control of a maneuver, such as an EOAM (Furda & Vlacic, 2011). Such primary systems are illustrated in Figure 12.

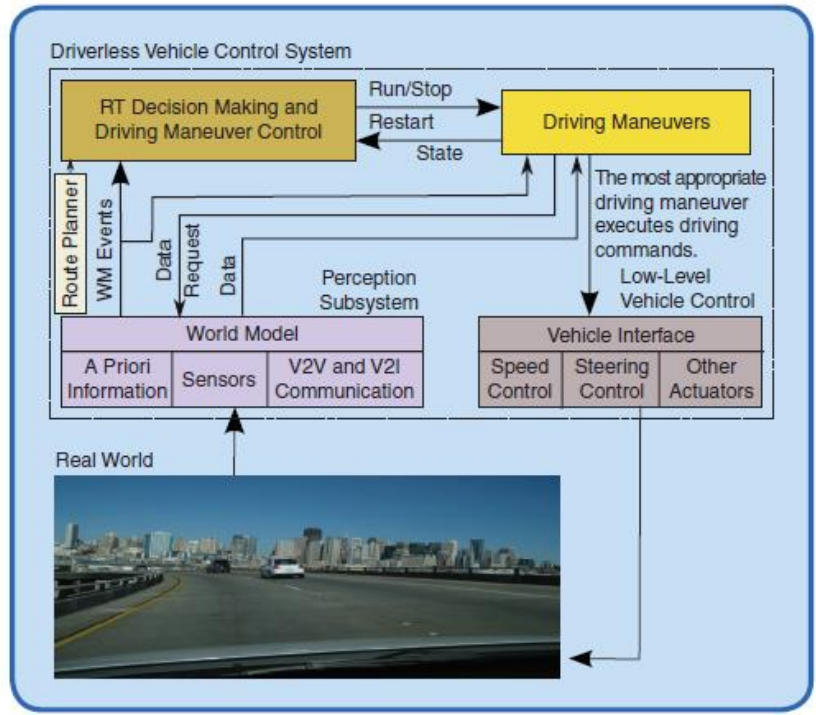

*Figure 12: ARV decision-making and control software architecture flow offered by (Furda & Vlacic, 2011)*

In a foundation work by (Amidi & Thorpe, 1990), overall questions that must be answered within a high-level system architecture of an ARV were posed as the following: Where am I? Where do I want to go? How do I get there without hurting myself? (Amer, Zamzuri, Hudha, & Kadir, 2017). Within the framework provided by (Amidi & Thorpe, 1990), these questions can be summarized by the following ARV system-level categories: Where am I? – Sensing and Perception, Where do I want to go? – Planning, and How do I get there without hurting myself? – Control. These system-level categories were enumerated by (Amer, Zamzuri, Hudha, & Kadir, 2017), in the following way:

**Sensing and Perception** – To provide real time data to let the system know the real time location and environment around the vehicle and prepare the raw data in a format that is feasible for the system to process.

**Planning** – To use the data provided by Sensing & Perception to dictate the safe and feasible path or trajectory for the vehicle to follow.

**Control** – Contain control strategies to move the vehicle on the desired path. This includes actuator control of each sub-system

To provide context for these definitions. (Amer, Zamzuri, Hudha, & Kadir, 2017) also provided an illustrative representation of the categories in the form of a system hierarchy (Figure 13).

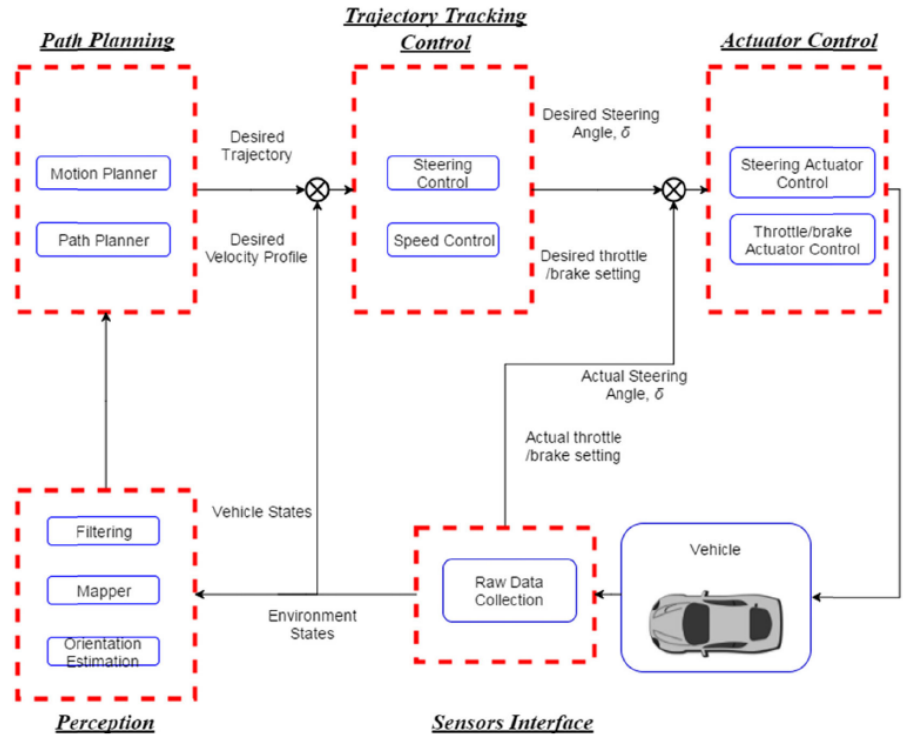

*Figure 13: System-level ARV hierarchy (Amer, Zamzuri, Hudha, & Kadir, 2017)*

In this hierarchy it is important to note that each high-level system has both inputs and outputs that demonstrate an entire closed loop (Amer, Zamzuri, Hudha, & Kadir, 2017). It is a common problem amongst other work, which define ARV system hierarchies that the high-level categories have outputs, without showing the inherent feedback of inputs that is necessary for each subsystem. Even the various vehicle sensors which obtain data based on their respective sensor function, also receive vehicle state information to help with exclusive calculations and fault detection (Furda & Vlacic, 2011) (Jo K. , Kim, Kim, Jang, & Sunwoo, 2014) (Jo K. , Kim, Kim, Jang, & Sunwoo, 2015).

A similar hierarchy that emphasized Perception, Planning, and Control as the primary systems, which was specifically applied toward a vehicle control software implementation was described by (Rodrigues, McGordon, & Gest, 2016) (Figure 14).

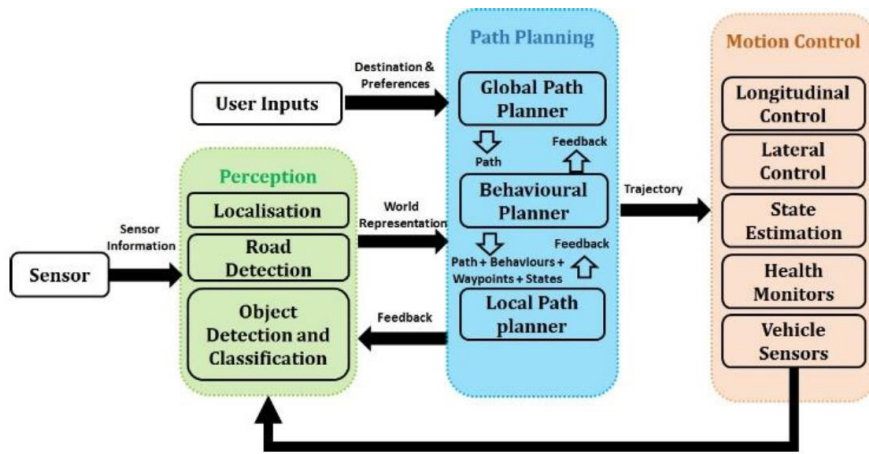

*Figure 14: Autonomous control software architecture defined in a system hierarchy format (Rodrigues, McGordon, & Gest, 2016)*

Comparable to (Amer, Zamzuri, Hudha, & Kadir, 2017), (Rodrigues, McGordon, & Gest, 2016) defines the Perception system as that which fuses sensed information of the vehicle surroundings to form a world representation; Path Planning as that which generates a future path from the vehicles current location to its intended destination; and Motion Control as the system which executes the planned path to reach the intended destination.

As a key part of the Sensing and Perception systems, (Furda & Vlacic, 2011) characterize the World Model as a software component which collects real-time status information from subsystems, while maintaining an up-to-date view of the vehicle's environment through sensor data, any available V2X[15] data, as well known a priori information[16] (Figure 12). The specific and relevant ARV sensor data can include wheel speed sensors, accelerometers, yaw rate sensors, global positioning systems (GPS) mapping, inertial navigation systems (INS), and inertial measurement units (IMU)[17]. The a priori data can also be comprised of any available V2X data, as well as road, intersection, and traffic data available before the ARV drive ever begins (Furda & Vlacic, 2011). Additionally, the as-described World Model provides access to all its information to other software components, systems and subsystems in a closed-loop manner, within the described hierarchy through an API (Application Programming Interface) (Furda & Vlacic, 2010).

In a similar representation as (Furda & Vlacic, 2010), (Claussmann, Revilloud, Gruyer, & Glaser, 2019) provide a representation of an ARV system hierarchy, that has a fundamental feedback loop between the low-level Actuation subsystem, and the high-level Motion Strategy system[18] (Figure 15).

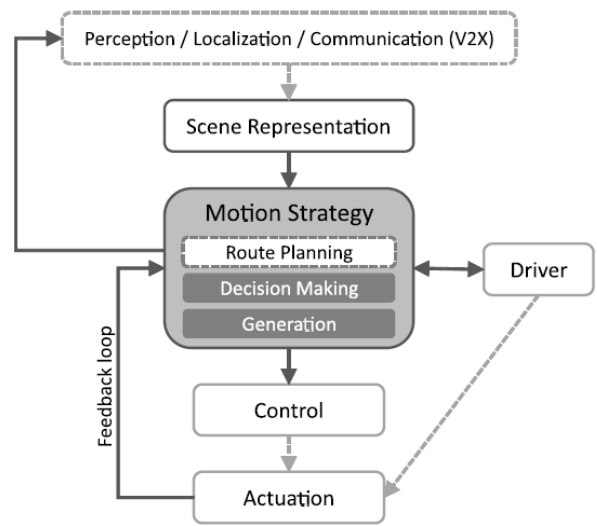

*Figure 15: Hierarchical system representation of an autonomous ground vehicle (Claussmann, Revilloud, Gruyer, & Glaser, 2019)*

Further detail of the Motion Strategy system that shows the interactions between the Route Planning, Prediction, Decision-Making, Generation, and Deformation subsystems were offered by (Claussmann, Revilloud, Gruyer, & Glaser, 2019), in Figure 16.

---

[15] V2X is commonly known as vehicle-to-anything (communication), with inclusions such as vehicle-to-vehicle (V2V) and vehicle-to-infrastructure (V2I)wireless communication methods

[16] (Furda & Vlacic, 2011) define a prior information as road, intersection, and traffic data which can be loaded before vehicle operation begins.

[17] It is noted that when IMU data are combined with GPS and/or INS data, the ARV can establish local heading and slideslip angles, which can be useful in EOAM trajectory generation and control.

[18] While there is a Driver block included in this system hierarchy, it is noted that the block would not exist for a fully autonomous (Level 5) (Society of Automotive Engineers, 2016) vehicle, but rather a vehicle that could allow for some intervention by a human driver.

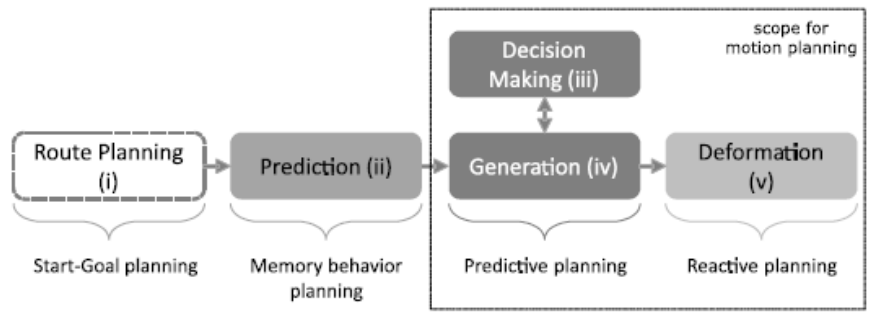

*Figure 16: Detailed view of the relevant subsystems within the Motion Strategy system (Claussmann, Revilloud, Gruyer, & Glaser, 2019)*

In this diagram representing details of the Motion Strategy system, (Claussmann, Revilloud, Gruyer, & Glaser, 2019) illustrated the following: the Route Planning subsystem maintains the global travel plan from start point to final destination; the Prediction subsystem compares the current vehicle state to historic dynamics data to predict the necessary future vehicle dynamics; the Decision Making provides a desired action for the vehicle based on input dynamics predictions; and the Generation subsystem creates a path which will effectively execute the selected motion decision. The Deformation subsystem completes reactive planning for instances that lie outside of those which can be predicted by the Prediction subsystem. An EOAM would fall within Deformation subsystem as described by (Claussmann, Revilloud, Gruyer, & Glaser, 2019).

A representative flowchart that fundamentally characterizes an ARV system-level hierarchy, can be seen in Figure 17 (Katrakazas, Quddus, Chen, & Deka, 2015).

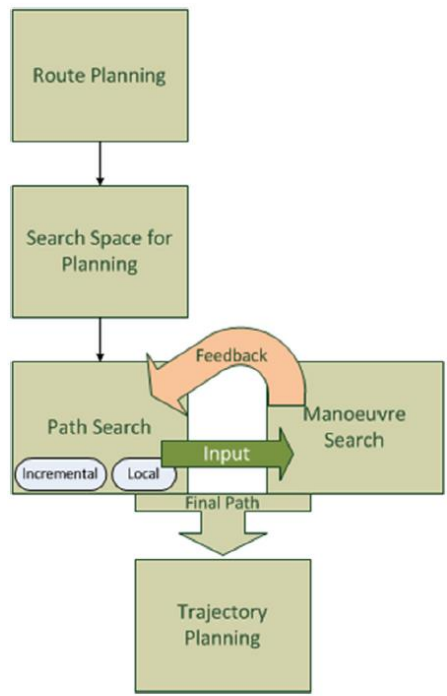

**Fig. 2.** A flow chart of planning modules.

*Figure 17: A flow chart representation of a system-level ARV hierarchy (Katrakazas, Quddus, Chen, & Deka, 2015)*

The four ARV hierarchical classes shown in Figure 17 suggested in an earlier work by (Varaiya, 1993) to be: (1) route planning, (2) path planning, (3) manoeuvre choice and (4) trajectory planning[19] (Katrakazas, Quddus, Chen, & Deka, 2015). Within the scope of an EOAM, the Manoeuvre Search and Trajectory Planning systems are the most closely related ones. Similar to the hierarchy defined by (Amer, Zamzuri, Hudha, & Kadir, 2017), this flow defined by (Katrakazas, Quddus, Chen, & Deka, 2015) show feedback that must exist for the real-time character of the ARV to successfully operate.

This hierarchy was extended by (Katrakazas, Quddus, Chen, & Deka, 2015) into a motion planning tree, as shown in Figure 18.

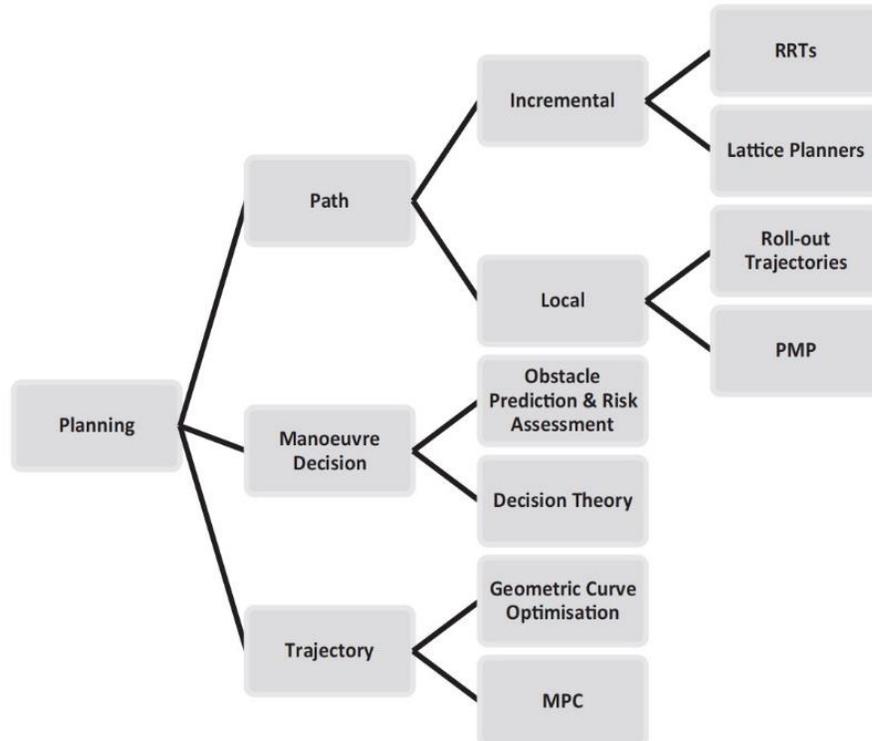

*Figure 18: Katrakazas summary of motion planning (Katrakazas, Quddus, Chen, & Deka, 2015)*

With this tree, it can be more clearly seen that the Manoeuvre Decision and Trajectory (Planning) subsystems yield low-level systems that directly form those utilized in an EOAM; for Manoeuvre Decision: Obstacle Prediction & Risk Assessment, and for Trajectory (Planning): Geometric Curve Optimisation and Model Predictive Control (MPC)[20].

## Obstacle Motion Prediction and Risk

Though they are not discussed in detail within this review manuscript, it is worth noting that the Obstacle Prediction & Risk Assessment categories highlighted by (Katrakazas, Quddus, Chen, & Deka, 2015), are relevant and advanced topics, with respect to an EOAM. The review by (Katrakazas, Quddus, Chen, & Deka, 2015) described Obstacle Prediction & Risk Assessment as subsystems which involve decision theoretic approaches, like those addressed in Markov Decision Processes (MDPs) and Game Theory, which are most often leveraged to account for vehicle and pedestrian interactions within a traffic environment (Katrakazas, Quddus, Chen, & Deka, 2015). In this area, Schwarting et al. 2018 mention various methods of probabilistic determination of obstacle, vehicle, and pedestrian motion intent (Schwarting, Alonso-Mora, & Rus, 2018). Meanwhile, Coskun focused on the MDP for lane changes in a highway environment (Coskun S. , 2018), and Sadigh utilized inverse reinforcement learning[21] (IRL) which utilizes continuous inverse optimal control (Sadigh, Sastry, Seshia, & Dragan, 2016). While worthy and certainly beneficial topics to aid in the advancement of EOAM strategies, these methods in predictive analysis of interactive elements of a traffic environment are outside of the scope of this manuscript and are left to the reader to explore in further detail.

---

[19] Note Varaiya 1993 originally labeled the lowest level system in the hierarchy to be Control Planning, but Katrakazas et al., 2015 updated this to be Trajectory Planning, within the context of their review (Katrakazas, Quddus, Chen, & Deka, 2015) (Varaiya, 1993)

[20] It should also be noted the Path system and associated subsystems in the tree provided by Katrakazas et al., 2015, represent those found in the General ARV Path Planning section this manuscript.

There are several other relevant and well-composed ARV systems hierarchies which have been drafted to explain the flow of logic and management of data within various ARV designs. While there are too many to cover in this manuscript, the reader is welcomed to explore the system hierarchies crafted by (Thrun, et al., 2006), (Jo K. , Kim, Kim, Jang, & Sunwoo, 2014) (Paden, Čáp, Yong, Yershov, & Frazzoli, 2016), and (Sharma, Sahoo, & Puhan, 2019), and (Peng, et al., 2020) among others.

## Actuator and Vehicle Modeling

Within the areas of ARV path and trajectory tracking (general, emergency maneuvers and others), vehicle modeling plays a key role in the ARV logic and control software. These vehicle models are used especially when tracking a reference path or trajectory that has been generated and can also be used when generating the paths or trajectories themselves (Katrakazas, Quddus, Chen, & Deka, 2015). The level of complexity of an ARV model is directly related to the number of model degrees of freedom (DOF), mathematical rigor needed to solve the relevant model differential equations (Claussmann, Revilloud, Gruyer, & Glaser, 2019).

Due to the Non-Holonomic and Kinodynamic nature of an ARV, the first sub-section of this Actuator and Vehicle Modeling section will be the Kinodynamic Considerations for an ARV. This will be followed by metrics for consideration in vehicle modeling constraints, latencies, and actuator dynamics, including those driven by Humans, Robot Actuators, and Electronic By-Wire systems. The final subsection will summarize various vehicle model types, which are commonly used in the literature.

### Kinodynamic Considerations

With respect to highway driving, (Claussmann, Carvalho, & Schildbach, 2015) mention the necessity for hard constraints – safety constraints of the environmental driving/traffic rules, to avoid collision, and soft (optimization) constraints – minimization of time, distance, or energy consumption and maximization of comfort (Claussmann, Revilloud, Gruyer, & Glaser, 2019). They, then, explained that hard constraints are essential and, soft constraints have room for relaxation. Kinodynamic considerations are other constraints, which focus on feasibility of motion. Claussmann et al., 2019 highlighted that an example of a kinematic constraint for an ARV is that it needs to travel on a smooth path which means a trajectory which is differentiable and has continuous curvature (Claussmann, Revilloud, Gruyer, & Glaser, 2019) while vehicle dynamic limitations related to steering, acceleration, braking, and tire/road friction are considered (Katrakazas, Quddus, Chen, & Deka, 2015). The term kinodynamic resides at the intersection of a vehicle's dynamic constraints, and its path-related kinematic constraints that directly result from the vehicle's inherent non-holonomic nature,

While holonomic and non-holonomic principals were explored early by (Whittaker, 1904), Amer et al., 2017, describes holonomic properties of a system as the relationship between the controllable degrees-of-freedom (DOF) and total number of DOF for the system (Amer, Zamzuri, Hudha, & Kadir, 2017). It was clarified by (Katrakazas, Quddus, Chen, & Deka, 2015) that the difference between holonomic robots and non-holonomic ARVs is that a robot is holonomic if the controllable degrees of freedom are equal to the total degrees of freedom. Alternatively, ARVs or car-like robots are non-holonomic because they are described by 4 DOFs: two Cartesian coordinates (x and y-directions), orientation movement (forward and backward) and heading angle, but have 2 kinematic constraints – (i)

---

[21] This method involves a reward function in which the ego ARV elicits behavior in vehicle operated by other drivers by purposefully changing the behavior of the human driver to either get that vehicle to slow down or reach its goal faster

they can only move backwards and forwards, tangentially to the direction of their main body and (ii) the steering radius is bounded (Katrakazas, Quddus, Chen, & Deka, 2015). Since there are only 2 controllable DOFs for an ARV, the number of controllable DOFs are less than the total system DOFs, and thus an ARV is non-holonomic.

It was also mentioned by (Elbanhawi & Simic, 2014) that kinodynamic planning includes constraints which are non-holonomic, kinematic, and dynamic. Likewise, Claussmann et al., 2019, highlighted that an example of a kinematic constraint for an ARV is that it needs to travel on a smooth path – a trajectory which is differentiable and has continuous curvature, and (Claussmann, Revilloud, Gruyer, & Glaser, 2019). Alternatively, some typical constraints related to the dynamics of the vehicle are those related to lateral handling (lateral acceleration, yaw rate, sideslip angle) (Choi, Kang, & Lee, 2012), (Manning & Crolla, 2007), longitudinal acceleration (power under traction and braking), as well as tire/road friction (Wang, Steiber, & Surampudi, 2008).

*Human-Driven Metrics and Constraints*

Additional characteristics of vehicle motion which should be considered in modeling ARVs, are latencies and constraints involved with control of the vehicle, and physical actuators used to control the ARV's motion. While physical ARVs utilize electronic control units (ECUs) with control logic embedded in software for their functions and controls, early and most current vehicle driving controls are managed by human drivers[22]. When considering a framework for designing ARV control constraints and latencies, those constraints and latencies involved with the human-vehicle system during various avoidance maneuvers can provide useful nominal baseline metrics.

A suitable method to understand these constraints within a natural driving environment that includes the driver-vehicle feedback control system, would be to conduct experiments which involve human drivers in passenger vehicles. One particularly influential study that involved a human driving analysis experiment relating to obstacle avoidance was that by (Maeda, Irie, Hidaka, & Nishimura, 1977). In this experiment, (Maeda, Irie, Hidaka, & Nishimura, 1977) conducted a participant study with a real vehicle on a test course, in which they would drive at a designated speed and attempt to avoid obstacles that appeared from behind barriers, laterally into the roadway[23]. The resulting maneuver was essentially a single lane change, and the driver was instructed to use steering only (Figure 19).

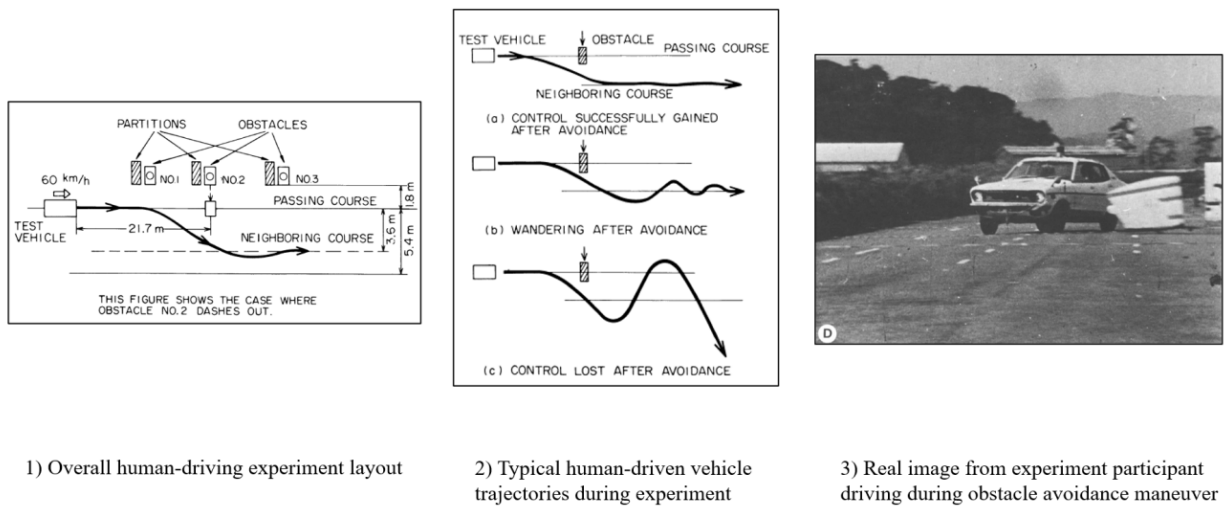

1) Overall human-driving experiment layout      2) Typical human-driven vehicle trajectories during experiment      3) Real image from experiment participant driving during obstacle avoidance maneuver

*Figure 19: 1) Overall experiment layout, 2) typical EOAM driver-vehicle outputs 3) Real experiment image capture from the (Maeda, Irie, Hidaka, & Nishimura, 1977) study*

---

[22] This excludes the passive and active safety systems discussed in the History of Safety Technologies section.
[23] This study also included a mechanism that each participant wore on their heads, which tracked their eye movement.

One primary focus of this study was to compare the vehicle control capabilities of both experienced and inexperienced drivers, during an obstacle avoidance maneuver. Another goal was to obtain human modeling parameter values from transfer function based on a (driver) model that was first proposed by (McRuer & Jex, 1967)[24].

With respect to identifying human performance characteristics, Maeda et al., 1977 found the following general performance metrics given in Table 1 to be true from his participant study.

*Table 1: A summary of human driving performance values during an EOAM experiment (Maeda, Irie, Hidaka, & Nishimura, 1977)*

| Human Driver Performance Metric | Maeda et al., 1977 |
|---|---|
| Test Speed [kph] | 60 |
| Steering Cycle Time (peak-to-peak) [s] | 2-2.5 |
| Driver Initial Sight Response Delay [s] | 0.4-0.5 |
| Driver Steer Response Delay (after Sight Response Delay) [s] | 0.3-0.4 |
| Start of Driver Maneuver (after Steer Response Delay) [s] | 0.1 |
| Total Delay Time [s] | 0.8-1.0 |
| Max Initial Steer Input [deg] | 200-230 |
| Max Initial Steer Rate Input [deg/s] | 700-900 |
| Max Yaw Rate Time Delay (before maneuver control difficulties) [s] | 0.3 |

With regard to these values (Table 1), it should be noted that the more experienced drivers had lower steering input magnitudes than inexperienced drivers (i.e., experienced drivers provided less input than inexperienced drivers)—the initial sight response was not affected by driver experience, and—the steer response delay was the time needed for the driver to adjust sight to the obstacle ahead (Maeda, Irie, Hidaka, & Nishimura, 1977). In their experiment, Maeda et al., 1970 noted that many drivers exhibited target fixation[25] with the obstacle ahead, which would not be an issue with an ARV with its sensors replacing human sight. It was also noted that an ARV or vehicle with robotic actuators could provide more accurate and precise electronic steering inputs, at faster steering rates than are capable by humans (Hingwe & Tomizuka, 1997) (Heydinger, 2005).

In 1980, (Reid, Graf, & Billing, 1980) utilized the modeling approach from (Maeda, Irie, Hidaka, & Nishimura, 1977) as a baseline for a driver model, with updates tailored toward a driving simulator study with participants. This study involved a simulator driving scenario on a straight two-lane road with no oncoming traffic, and an obstacle (falling pole) that appeared such that it blocked the lane in which the ego vehicle is initially traveling. As the obstacle entered the lane, the study participant was tasked with avoiding the object using steering only with a least single lane change, though a double lane change was allowed (Reid, Graf, & Billing, 1980) (Figure 20).

---

[24] Like McRuer's model from 1957, this Quasi-Linear Pilot model was initially intended for combat aircraft pilot modeling

[25] When eyes are known to fixate on an object until that object is outside of the driver's field-of-view (FOV)

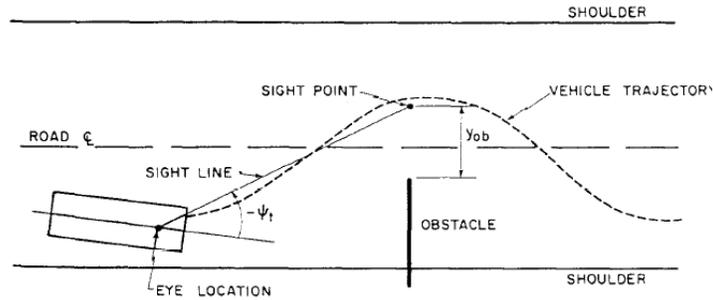

*Figure 20: Simulator study EOAM (Reid, Graf, & Billing, 1980)*

In this study, Reid et al., 1980 found that the max steer rate was 16 rad/s (917 deg/s), which agrees well with that found by Maeda et al., 1977.

Another human driver participant study on a driving simulator, with a focus on obstacle avoidance was conducted by (Soudbakhsh, Eskandarian, & Moreau, 2011). The objective of that research was to develop a steering control collision avoidance system, which takes the driver's steering input to trigger the system to perform a fast and successful avoidance maneuver, in which the internal control logic conducts the maneuver, including actuator inputs. The vehicle speed was 25 m/s, on a two-lane highway of 4,600m in length, having one lane one each side, with both rural and urban surroundings.

The avoidance maneuver was largely based on that developed by (Chee & Tomizuka, 1994), centered on a trapezoidal acceleration profile, with similar constraints for maximum lateral acceleration and jerk (utilized for comfort). If, in the study, no steering was initiated by the required time, braking was initiated automatically to mitigate imminent collision. The general decision-making logic flow for this maneuver can be seen below (Figure 21).

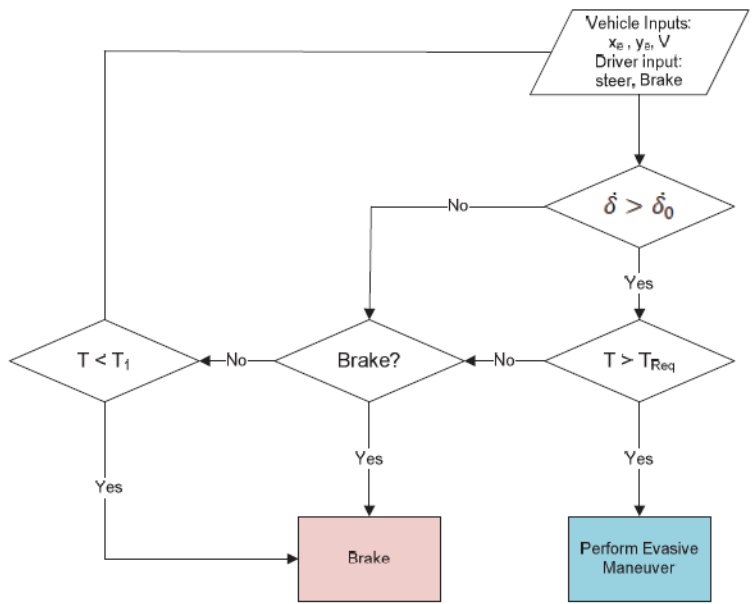

*Figure 21: Flow chart showing algorithm for a human-driver initiated obstacle avoidance maneuver (Soudbakhsh, Eskandarian, & Moreau, 2011)*

From the experiment, (Soudbakhsh, Eskandarian, & Moreau, 2011) found that the participants' average initiation of the necessary steering input to trigger the automatic maneuver was at a time-to-collision[26] of 1.21s, while duration of the maneuver averaged to be 1.13s, and the driver's average reaction time was 0.7s to recognize the object and steer initially (Soudbakhsh, Eskandarian, & Moreau, 2011). A summary of relevant human response characteristics related to EOAMs for (Maeda, Irie, Hidaka, & Nishimura, 1977), (Soudbakhsh, Eskandarian, & Moreau, 2011), and (Reid, Graf, & Billing, 1980) can be found in Table 2.

*Table 2: Summary of Human Driver Performance Metrics for EOAMs, from Maeda et al., 1977, Reid et al., 1981, and Soudbkhash et al., 2011*

| Human Driver Performance Metric | Maeda et al., 1977 | Reid et al., 1981 | Soudbkhash et al., 2011 |
|---|---|---|---|
| Test Speed [kph] | 60 | 60 | 90 |
| Steering Cycle Time (peak-to-peak) [s] | 2-2.5 | - | 1.13 (avg) |
| Driver Initial Sight Response Delay [s] | 0.4-0.5 | - | - |
| Driver Steer Response Delay (after Sight Response Delay) [s] | 0.3-0.4 | - | - |
| Start of Driver Maneuver (after Steer Response Delay) [s] | 0.1 | - | - |
| Total Delay Time [s] | 0.8-1.0 | - | 0.7 |
| Max Initial Steer Input [deg] | 200-230 | - | - |
| Max Initial Steer Rate Input [deg/s] | 700-900 | 917 (avg) | 700-900 |
| Max Yaw Rate Time Delay (before control difficulties) [s] | 0.3 | | - |

## Robot Actuator-Driven Metrics and Constraints

While the human-driving performance tendencies, capabilities, and constraints during an EOAM serve as good references for target metrics of an ARV, potentially more realistic representations of input capabilities and constraints for an ARV can be seen with robot actuators designed for steering, throttle, and braking. In this section, we will be focusing on robotic steering actuators, and their performance capabilities and constraints with respect to EOAM studies.

One prominent study using an early robotic steering actuator was that by (Tseng, et al., 2005), where an Anthony Best Dynamics (ABD) steering robot was used to conduct automated and closed-loop steering tasks for EOAMs. Specifically, the ABD SR30 robot[27] was used, which utilizes a brushless motor with a mass of 10kg which has max torque of 35 Nm and a rated torque of 30 Nm at 1000°/s, > 10Nm at 2000°/s (Industrial Measurement Solutions, 2020).

---

[26] It is noted that time-to-collision is a well-documented metric for risk of collision with an ARV. Some relevant references for this metric can be found in the following works: ( (Haus, Sherony, & Gabler, 2019) (Hamberg, Hendriks, & Bijlsma, 2015) (Lefèvre, Vasquez, & Laugier, 2014) ).

[27] Note that at the time of this manuscript the SR30 steering robot has been replaced with the SR35 (Anthony Best Dynamics, 2020)

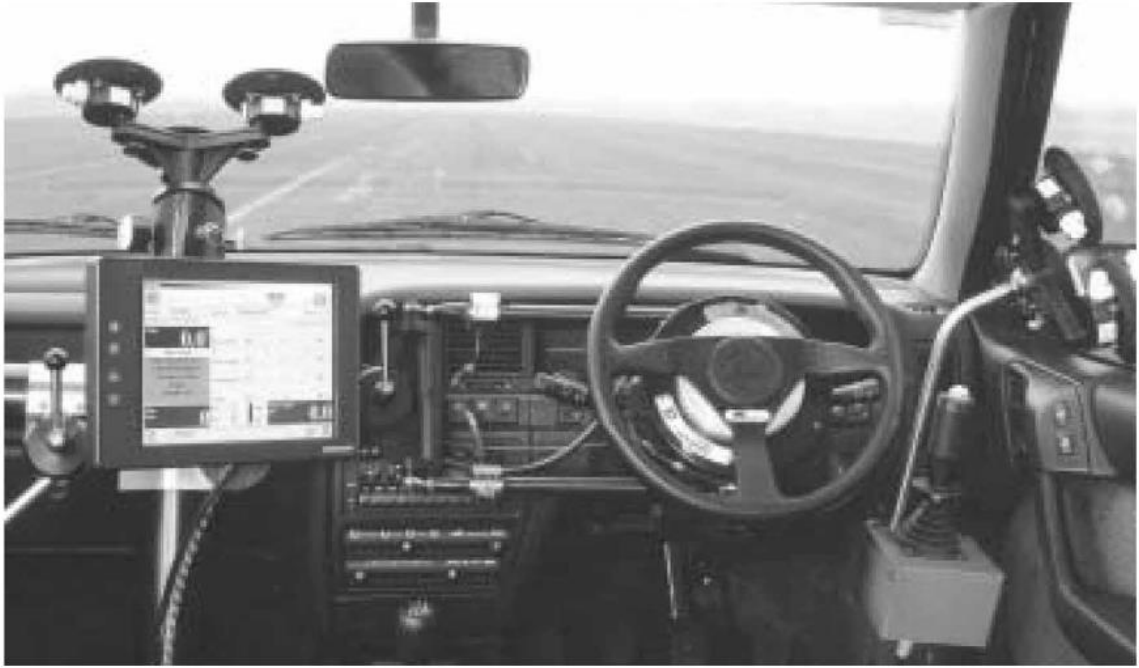

*Figure 22:ABD SR30 Steering robot and controller setup for automated evasive maneuver test (Tseng, et al., 2005)*

Clearly these capabilities of the steering robot vastly exceed those of a human operator (Table 1) in both torque, and steering rate, and could be considered as the new baseline for steering actuator constraints in an ARV. The steering driver model used single point preview with lateral path error as feedback, based on (McRuer & Weir, 1969), (McLean & Hoffmann, 1973), (McRuer, Allen, Weir, & Klein, 1977). The lateral control and trajectory-tracking utilized parameter space robust control by (Ackermann J. , 1980), and more recently utilized Sliding Mode Control (SMC) that was modernized by (Hingwe & Tomizuka, 1997). One important note about SMC when used with robotic steering actuators is that delays and imperfections in physical actuators, use of sliding mode control can cause "serious actuator and plant damages, energy losses, and unwanted disturbance due to" chattering (Khalil & Grizzle, 2002) (Amer, Zamzuri, Hudha, & Kadir, 2017). The main takeaway, however, is that the torque and steering rate of the steering robot could represent a new set of physical actuator constraints, for consideration in EOAM modeling. Some sample output from the study by (Tseng, et al., 2005), which shows the repeatability of the steering robot when conducting several double lane change (DLC) maneuvers at 15 m/s (54 kph) can be seen in the lateral acceleration vs. time plot in Figure 23.

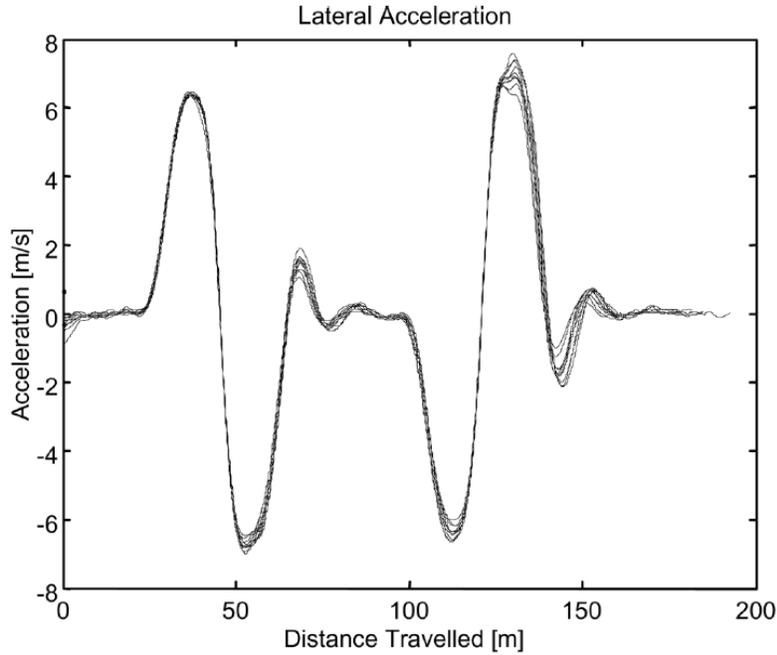

Figure 10. Focus repeatability – – lateral acceleration.

*Figure 23: Driving robot lateral acceleration for DLC, repeatability result (Tseng, et al., 2005)*

Experiments conducted by (Heydinger, 2005) and Mikesell (Mikesell, Sidhu, Guenther, Heydinger, & Bixel, 2007) also incorporated a steering robot, though the manufacturer was different. First (Heydinger, 2005), then later (Mikesell, Sidhu, Guenther, Heydinger, & Bixel, 2007) utilized the SEA – automated steering controller (ASC)[28], which offered max torque of 75 Nm, a rated torque of 69 Nm at 720°/s, and maximum angular rate of 1800°/s, also using a brushless motor with a mass of 14.5kg (SEA Limited, 2020a)(Figure 24).

---

[28] Note that at the time of this manuscript the ASC steering robot has been updated to the version that is used with SEA's automated test driver (ATD), which is lighter and has slightly different output capabilities  (SEA Limited, 2020b)

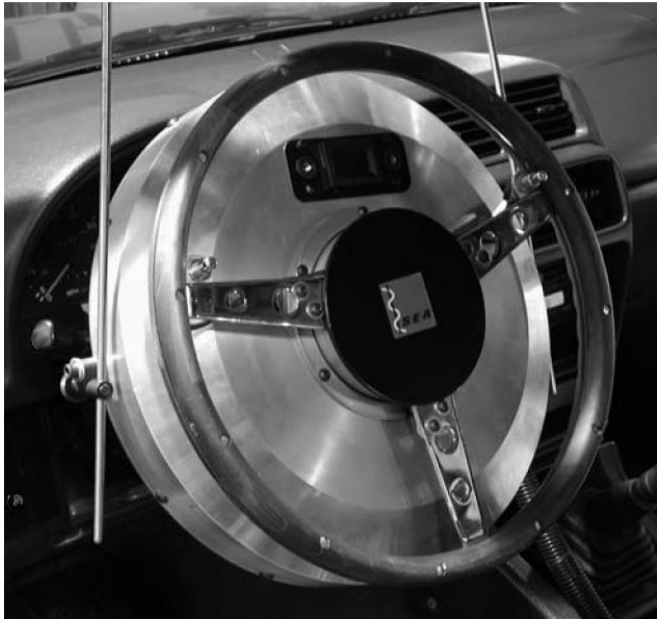

*Figure 24: SEA ASC steering motor assembly, installed in a test vehicle (SEA Limited, 2020a)*

Key performance outputs of the ABD and SEA robotic steering motor assemblies can be seen in Table 3.

*Table 3: Comparison of robotic steering actuator performance values (Tseng, et al., 2005), (Heydinger, 2005)*

| Human Driver Performance Metric | Tseng et. al 2005 | Heydinger 2005 |
|---|---|---|
| Motor Type | brushless | brushless |
| Mass [kg] | 14.5 | 14.5 |
| Peak Torque [Nm] | 35 | 75 |
| Rated (nominal) Torque [Nm] at Rotation Rate [°/s] | 30 at 1000 | 69 at 720 |
| Peak Angular Rate [°/s] | 2000 | 1800 |

While the steering robot systems above provide a vast expansion in capabilities and broadening of constraints for steering inputs which could be applied for an ARV conducting an EOAM, in the state-of-the art, column, and rack-mounted electronic motors are able to control steering without adding extra components. These are often the motors used for modern ARVs because the designed motors are well-integrated into the vehicle, and only require appropriate controllers and access to internal vehicle communication networks[29].

*Electronic By-Wire-Driven Constraints*
In many modern vehicles there are at least electronically assisted power steering systems (Kozaki, Hirose, Sekiya, & Miyaura, 1999) (Sulakhe, Ghodeswar, & Gite, 2013), and completely electronic throttles (McKay, Nichols, & Schreurs, 2000).

---

[29] Modern day automotive vehicle communication networks are CAN, Ethernet, and FlexRay (Hank, Müller, Vermesan, & Van Den Keybus, 2013)

*Electronic Throttle Application*

While electronic throttles[30] have been a crucially important feature in vehicle powertrain development, they have been included on most new passenger vehicles for over 20 years (at the time of this manuscript) (McKay, Nichols, & Schreurs, 2000). Due to the current commonplace of electronic throttles in modern passenger vehicles, details regarding electronic throttles will be left to reader to explore.

*Electronic-Assisted Steering*

One example of the electronic-assisted steering system that is heavily involved in EOAMs is electronic power steering (EPS) (Japan Patent No. JPS59130780A, 1983). These systems can have electronic assist on the steering column, rack, pinion, or direct drive gear (Clemson University Vehicular Electronics Laboratory, 2021) Like the robotic steering actuators mentioned above, characteristics of EPS, largely depend on the motor attached, with up to 10 Nm or steering column torque and up to nearly 14 kN of steering rack force available (Robert Bosch GmbH, 2020a). In an automated vehicle tested by Ryu et al., 2013, the electronic steering controller issued commands at 100 Hz, and the column-mounted motor was capable of 5.8 Nm of continuous torque at 4000 RPM (66.6 rotations/s, 418.9 rad/s), from a 24V battery (Ryu, Ogay, Bulavintsev, Kim, & Park, 2013); at this speed, the column-mounted steering motor can rotate at nearly 24,000 deg/s. These electronically-assisted steering rates exceed those of a human driver, even when considering associated steering system gear ratios when turning a steering wheel, versus those of a column-mounted or rack mounted electronic-assisted steering system (Maeda, Irie, Hidaka, & Nishimura, 1977).

*Electronic-Assisted Braking*

While some hybrid (Li, et al., 2016) (Gao & Ehsani, 2001) (Nadeau, Micheau, & Boisvert, 2017) and electric vehicles (Beji & Bestaoui, 2005) have regenerative braking in combination with conventional hydraulic brakes, full by-wire braking systems have not yet been adopted, outside of certain test prototypes, like those used in the Stanford Dynamic Design Laboratory (DDL) (Yih & Gerdes, 2005) (Laws, et al., 2005) (Funke, et al., 2012) (Goh & Gerdes, 2016) (Goh, JY; Goel, T; Gerdes, CJ, 2020).

Also used in DDL were test vehicles that utilize active brake boosters (Talvala, Kritayakirana, & Gerdes, 2011) (Kritayakirana & Gerdes, 2012) (Funke & Gerdes, 2016) (Laurense, Goh, & Gerdes, 2017) (Subosits & Gerdes, 2019). These vacuum-independent active brake boosters, and others that are similar, maintain a hydraulic connection from the brake pedal to the calipers, but utilize an electronic force-feedback mechanism (Robert Bosch GmbH, 2020b). This electromechanical brake system can provide the driver with both solid pedal feel in all conditions, and more precise brake boost control, than conventional purely mechanical brake boosters. By being electronic and vacuum independent, these electromechanical brakes allow for automated driving in both ADAS features like Automatic Emergency Braking (AEB), and level 3 or above autonomous driving, since such braking systems can electronically control the pedal, without driver intervention (Robert Bosch GmbH, 2020b). A demonstration of electro-hydraulic braking (EHB) system, like that made by Bosch, with both pressure build and release times from the experiments conducted were shown by (Han, Xiong, & Yu, 2019)(Figure 25 and Figure 26–units in seconds).

---

[30] Also known as throttle-by-wire

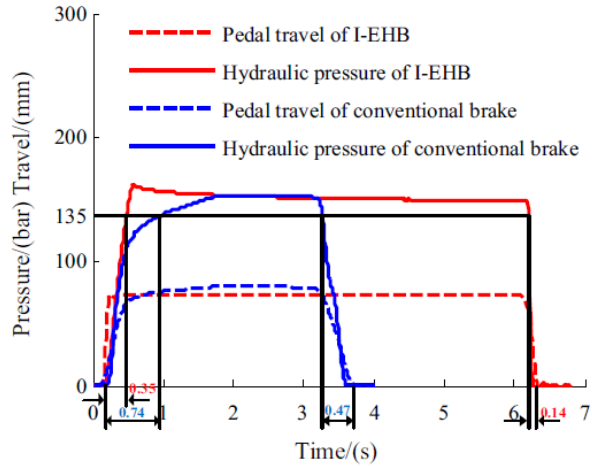

*Figure 25: Conventional and EHB system experimental timing comparisons (Han, Xiong, & Yu, 2019)*

|  | Pressure-rising (from 0 to 135 bar) | Pressure-reducing (from 135 to 0 bar) |
|---|---|---|
| EHB | 0.35 | 0.14 |
| Conventional brake system | 0.74 | 0.47 |

*Figure 26: Pressure build and release timing comparison between conventional and EHB–units in seconds (Han, Xiong, & Yu, 2019)*

These pressure build and release timings were found, using an electronic actuator, experimentally; however, they give an idea of system delay timings that can be expeted with simialr EHB systems, which can be used for ARV braking.

There have been certain vehicles that have systems which remove the mechanical connection from the brake pedal to the hydraulic braking unit, and these systems can be categorized as partially by-wire systems (American Honda Motor Co., Inc, 2012)[31].

---

[31] There is a valve bypass for this system that directly connects the brake pedal to the friction brakes, during an electronic failure

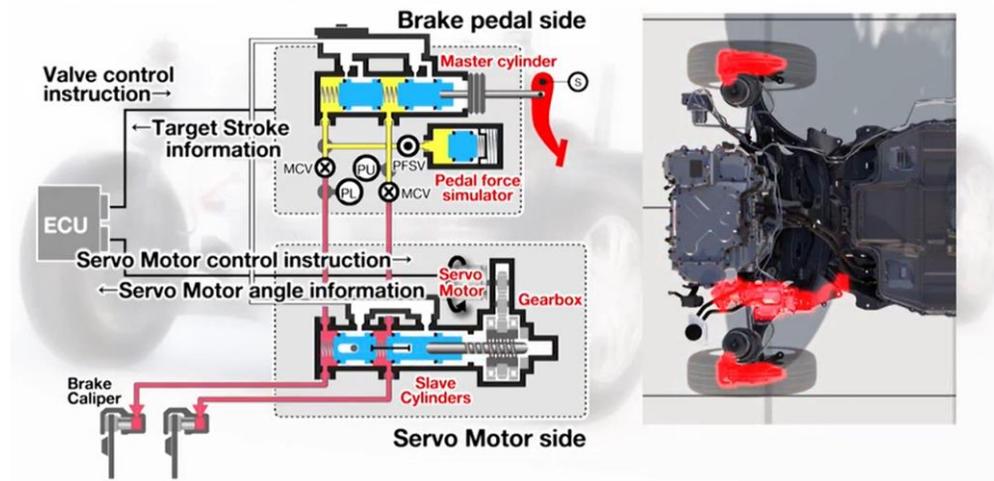

*Figure 27: Example of friction braking application with the Honda Motors Electro-Servo Braking (ESB)system (American Honda Motor Co., Inc, 2012)*

While it is expected that these systems have even quicker brake fluid pressure build and release, due to removing a mechanical connection between the pedal and the tandem master cylinder (Japan Patent No. US8897984B2, 2012) (unless a failsafe condition occurs[32]), values for this performance, are not readily available at the time this manuscript was written.

*Automatic Emergency Braking (AEB) Performance as related to ARVs*

Still, when looking at AEB and/or ARV braking one of the key metrics for modeling timing and tuning for autonomous braking logic and associated system latencies, is time-to-collision (TTC) (Haus, Sherony, & Gabler, 2019). In a 2019 study, (Haus, Sherony, & Gabler, 2019) took Unites States pedestrian crash data, and incorporated theoretical AEB systems[33] with those data, utilizing varying latency values and TTC initiation values for the AEB systems (essentially, different TTC tuning values for each system). The results of this study showed as AEB system latency decreased, and TTC increased, the percentage of pedestrian collision events that were avoided, also increased regardless of the type of braking applied (late and hard apply or early and weak apply) (Figure 28).

---

[32] A failsafe condition is noted as that when and electronic failure occurs, and a redundant backup system allows for safe operation
[33] These AEB systems were primarily focused on pedestrian detection

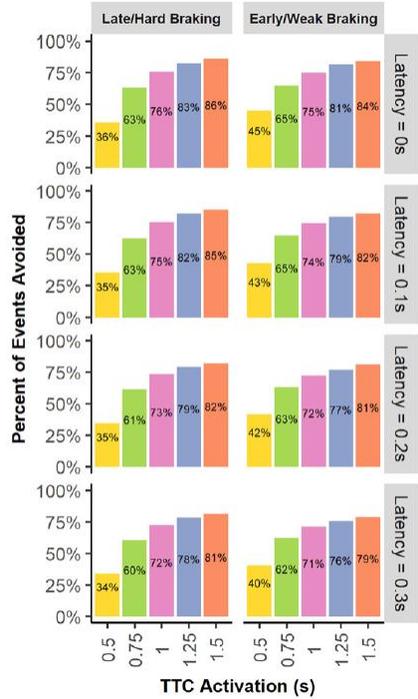

*Figure 28: Estimated percentage of pedestrian collision events avoided based on TTC and latency of theoretical AEB systems (Haus, Sherony, & Gabler, 2019)*

The same trends were also shown for percent reduction of fatality risk (Figure 29) (Haus, Sherony, & Gabler, 2019)

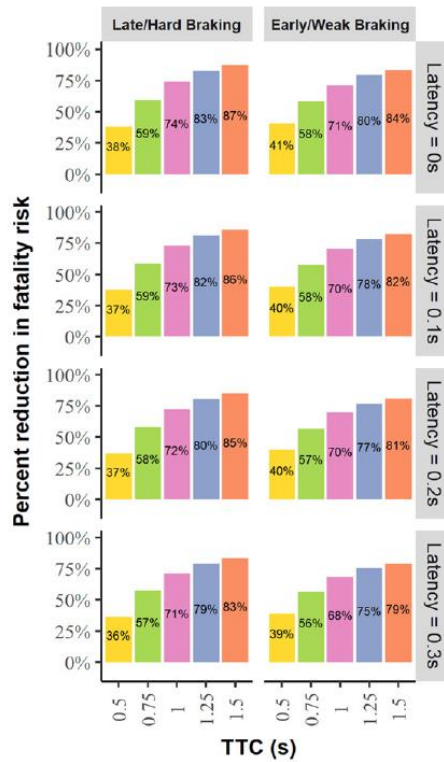

*Figure 29: Percent change in fatality risk of theoretical AEB systems applied to historical for vehicle-pedestrian crash data (Haus, Sherony, & Gabler, 2019)*

These statistics and more highlight the importance of pursuing low system latency for autonomous braking systems that can be utilized by ARVs, and also choosing appropriate TTC values for those AEB systems, without making them too high, which can cause them to be hypersensitive and lead to false positives (Edwards, Nathanson, & Wisch, 2014).

To further characterize an autonomous braking goal of at least matching human behavior in dense traffic, in Figure 30 (Hamberg, Hendriks, & Bijlsma, 2015) show various latencies involved with AEB, and that if there is a way to provide warning that a braking event will occur (i.e., for an intelligent vehicle-to-vehicle (V2V) wireless communication system, that warns following vehicles), the entire process of AEB can start earlier.

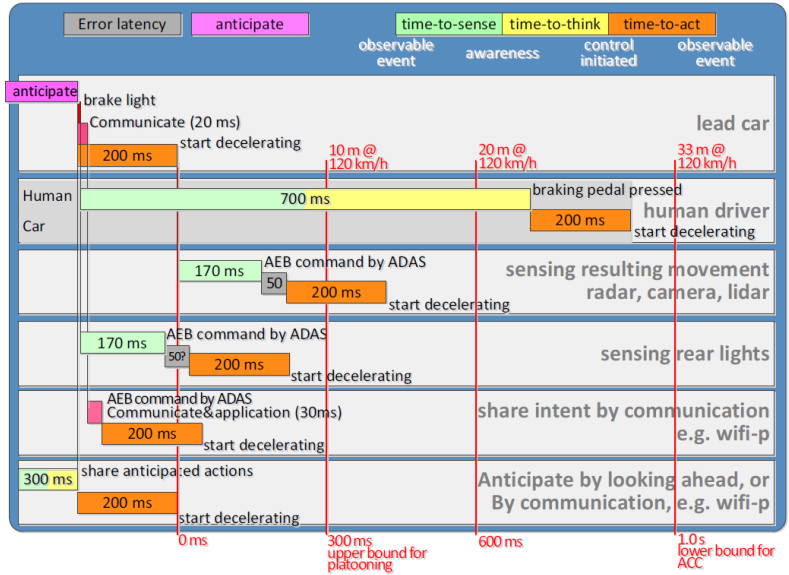

*Figure 30: Relevant AEB system for a lead vehicle, human driver, and ARV latencies with various levels of lead vehicle anticipation capabilities (Hamberg, Hendriks, & Bijlsma, 2015)*

In Figure 30, it is important to note that (Hamberg, Hendriks, & Bijlsma, 2015) report that when traveling at 120 km/h, the time during actual emergency braking is 200 ms, time to sense that a braking event is necessary is 170 ms, and some error involved in detection is about 50 ms, when braking autonomously. While the time to brake is shown to be 200 ms for human drivers, the time to sense a braking event and respond is much longer (700 ms total).

### ADAS and ARV Sensor Dynamics
It should be noted that many other studies have been completed on automotive steering, braking, and throttle actuator dynamics, along with latencies of various sensors used for vehicle, vulnerable road user (VRU), and general obstacle detection. While this information is important and evolving with the pace of new ARV and ADAS component development, it is out of the scope of this manuscript and should be investigated by the reader. Some resources on automotive steering, braking, and throttle actuator dynamics for the reader to explore, are the following: (Hedrick, Tomizuka, & Varaiya, 1994), (Tai & Tomizuka, 2000), (Fujiwara, Yoshii, & Adachi, 2002), (Line, Manzie, & Good, 2008), (Babu, et al., 2019), and (Theerthala, 2019), (Emirler, Uygan, Aksun-Guvenc, & Guvenc, 2014), (Oncu, et al., 2007), (Aksun-Guvenc, Guvenc, Ozturk, & Yigit, 2003), (Guvenc, et al., 2012), (Bowen, Gelbal, Aksun-Guvenc, & Guvenc, 2018).

### Vehicle Model Types
In a general sense, a vehicle model for ARV simulation must be capable of motion which is "feasible, safe, optimal, usable, adaptive, efficient, progressive, and interactive" (Rodrigues, McGordon, & Gest, 2016). Earlier, we mentioned that for ARVs the modeling is inherently non-holonomic, and kinodynamic. Again, non-holonomic systems are those which include non-integrable constraints, such that the number independent configuration variables exceed the available degrees of freedom (DOFs) (Whittaker, 1904), and kinodynamic modeling includes considerations with simultaneous kinematic (motion and trajectory) and dynamic (velocity, acceleration, force) constraints (Canny, Donald, Reif, & Xavier, 1988).

Such non-holonomic and kinodynamic vehicle models are primarily utilized to simulate vehicle responses to various inputs, when either creating a path, or constructing a controller/tracking algorithm (Amer, Zamzuri, Hudha, & Kadir, 2017), as well as even understanding understeer or oversteer handling characteristics ( (Milliken & Whitcomb, 1956), (Bundorf, 1968)). These non-holonomic and kinodynamic vehicle models are also often built into the trajectory generation and trajectory-tracking controller logic that is used for ARV software creation. More details about these vehicle models in the context of ARV EOAMs will be discussed below.

*Model Selection*

When selecting a vehicle model, one of the primary characterizations to consider is the number of DOFs that the model will have. The vehicle model and its DOFs affect both the motion planning and the control architecture for the motion (Claussmann, Revilloud, Gruyer, & Glaser, 2019). Some authors in the field also commented that the vehicle model and its DOFs should address ARV driving motion, safety (Schürmann, et al., 2017), consistency (Polack, Altché, d'Andréa-Novel, & de La Fortelle, 2018), as well as the evolution of the ego ARV in time (Claussmann, Revilloud, Gruyer, & Glaser, 2019).

Due to the ARV being highly nonlinear in its combined components and systems, careful choices regarding whether to model ARVs and their systems as linear or nonlinear is critical, as the model linearity (or lack thereof) will affect its complexity, computation speeds, and available solver tools. This choice of vehicle model DOFs often has to do with finding the minimum viable model which is suitable for the target application or experiment. For example, (Claussmann, Revilloud, Gruyer, & Glaser, 2019) mention that for highway driving with automated lane change and cruise control maneuvers, a vehicle model with high DOFs is not generally necessary due to the relatively simple lateral and longitudinal dynamics involved. This, of course, is considering mild nominal conditions of daylight, and mild ambient temperatures, no precipitation, and high surface friction coefficient.

*General Vehicle Model Groupings*

Broad categories of vehicle models have been succinctly arranged by (Amer, Zamzuri, Hudha, & Kadir, 2017) and can be seen in Figure 31 and Figure 32. The highest-level categories listed are Geometric, Kinematic, and Dynamic.

| Model Type | Vehicle Model | Strength | Weakness | Comment(s) |
|---|---|---|---|---|
| Geometric | Geometric model based on Ackermann steering configuration | Simple, less parameter needed | No indication on internal forces | Simple configuration and parameters |
| | | Sufficient to describe relationship between vehicle position and path. | | Suitable for purposes that does not require velocity and acceleration responses. Only vehicle position indication |
| | | Usually used to develop geometric controller | | |
| Kinematic | Full vehicle kinematic model | Take into account the possibility of different direction between vehicle heading relative to local coordinates, $\theta$ and global coordinates, $\psi$ | No indication on internal forces | Consider left and right wheel |
| | | | Slightly more complicated than geometric model | Suitable for purposes that does not require dynamics responses. |
| | Half (bicycle) vehicle kinematic model | Simple, less complex configurations | No indication on internal forces | Consider only one wheel per axle. |
| | | | Assume vehicle heading with respect to local coordinates is the same as heading with respect to global coordinates | Suitable for purposes that does not require dynamics responses. |
| | Kinematic model with slip angle | Consider slip in the model for manoeuvrings | No indication on internal forces | Suitable for studies on vehicles operating on slippery surfaces and high speed range |
| | | | Added complexity | |
| Dynamic (linear) | Full vehicle kinematic model | Consider forces in all wheels, especially in cornering manoeuvres | Does not consider non-linear function of tire properties | Suitable for studies where dynamic observation is important but small slip $\approx 0.5$g acceleration and $\approx 5$ slip angle |
| | Half vehicle kinematic model | Less complicated model | Does not consider non-linear function of tire properties | Same as above |
| | | | Neglect the effect of different responses for left and right tires | The difference of behaviour in right and left tire may become significant during cornering |

*Figure 31: Geometric and Kinematic vehicle model category characteristics (Amer, Zamzuri, Hudha, & Kadir, 2017)*

| Model Type | Vehicle Model | Strength | Weakness | Comment(s) |
|---|---|---|---|---|
| Dynamic (nonlinear) | Full vehicle kinematic model | Consider forces in all wheels, especially in cornering manoeuvres | More forces included. More complex | One of the complete handling model |
| | | Consider the nonlinearity of tire responses with respect to slip angle | | Suitable for studies on vehicle subject to different manoeuvrings |
| | Half vehicle kinematic model | Less complicated model | Does not consider the effect of inner and outer wheel during cornering | Suitable for studies on vehicle subject to different manoeuvrings |
| | | Consider the nonlinearity of tire responses with respect to slip angle | | |

*Figure 32: Dynamic vehicle model characteristics (Amer, Zamzuri, Hudha, & Kadir, 2017)*

In Figure 31 and Figure 32, each vehicle model is considered for its strengths, weaknesses, and other notable comments regarding the models' performance is presented. The Geometric vehicle model has a single characterization whereas the Kinematic has subcategories including a Full model (considering all four wheels and tires), Half model (left and right tires are combined to one[34]), and that which includes slip angle. The Dynamic model can also include subcategories of Full and Half that are characterized as both linear systems and nonlinear range systems, and often include either linear or even nonlinear tire models (Amer, Zamzuri, Hudha, & Kadir, 2017). Details regarding canonical Full and Half models commonly used in vehicle dynamics lateral handling models can be found in (Gillespie, 1992), (Genta, 1997), (Wong, 2008).

*Geometric and Kinematic Models*
The Geometric model's primary purpose is to help with a spatial understanding of the vehicle's dimensions and trajectory-tracking capabilities (Thrun, et al., 2006) (Amer, et al., 2018) with respect to the desired path geometry, but without including kinematic or dynamic details. The Kinematic model increases model complexity slightly by including position, velocity, and acceleration determinations without suggestion of internal forces due to mass. The Dynamic model builds on the Kinematic model by including internal forces, and can be represented in both linear and nonlinear model forms (Gillespie, 1992), (Genta, 1997), (Wong, 2008), (Amer, Zamzuri, Hudha, & Kadir, 2017).

Often in the literature, one vehicle model type is utilized for the trajectory generation, and another for the trajectory-tracking controller. Often, the model for the trajectory generation can be simplified, as only a kinematic representation is required in most cases. However, for the trajectory-tracking controller, higher dimensional complexity is often required so that motions of the virtual ARV may be representative as closely as possible of an actual vehicle, given the constraints of the processing hardware, and calculation time limits (Falcone, Borrelli, Tseng, H. E., & Hrovat, 2008) (Katrakazas, Quddus, Chen, & Deka, 2015) (Amer, et al., 2018), (Li, Zhu, Aksun-Guvenc, & Guvenc, 2021).

*Point Mass Model*
As the most basic vehicle model form, a point mass vehicle model is able capture the basic vehicle motion along a trajectory, utilizing simplified assumptions about the vehicle characteristics; the key advantages are that the point mass model is linear, and if it is dynamic, forces are applied directly, to drive the model motion. The linearity and direct force application of a point mass model allows an online optimization problem for specific trajectory, given various constraints and cost function to become analytically tractable (Shiller & Sundar, 1998), (Subosits & Gerdes, 2019). This straightforward analytical solution for the trajectory as a function of time for a point mass model lends it to be used for online/real-time trajectory generation (Shiller & Sundar, 1998), (Falcone, Borrelli, Tseng, H. E., & Hrovat, 2008) (Gao, Lin, Borrelli, Tseng, & Hrovat, 2010) (Funke & Gerdes, 2016) (Subosits & Gerdes, 2019), (Zhu & Aksun-Guvenc, Trajectory Planning of Autonomous Vehicles Based on Parameterized Control Optimization in Dynamic on-Road Environments, 2019) as it allows for the trajectory to be determined purely by kinematics (Homann, et al., 2017). As defined by (Amer, Zamzuri, Hudha, & Kadir, 2017), the point mass model would be categorized as a Dynamic (linear) vehicle model due to consideration of the model acceleration and internal forces, although those forces are concentrated to a single point.

*Kinematic vs. Dynamic Bicycle Model Selection*
As at least one quantifiable metric for model selection in ARV trajectory planning (or generation), (Polack, Altché, d'Andréa-Novel, & de La Fortelle, 2017) presented a criterion for determining appropriate conditions to use a 3 DOF kinematic bicycle model, and when a more complex model is best suited, especially when used with an MPC controller. This criterion described that the limit at which a kinematic bicycle model can be used is at $a_{y_{max}} = 0.5 \, \mu g$, or the lateral acceleration limit should be half the value of the available friction force between the tire and the road (Polack, Altché, d'Andréa-Novel, & de La Fortelle, 2017) (Polack, Altché, d'Andréa-Novel, & de La Fortelle, 2018).

Graphical representations of the bicycle model used by (Polack, Altché, d'Andréa-Novel, & de La Fortelle, 2017) for this criterion can be seen in Figure 33.

---

[34] This model where the left and right tires are combined into one for both the front and rear is often called the bicycle model

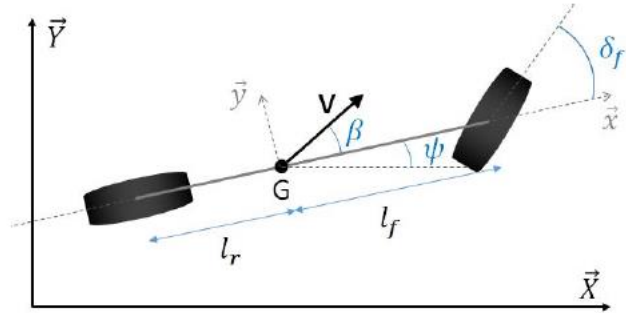

*Figure 33: Kinematic bicycle model as proposed by (Polack, Altché, d'Andréa-Novel, & de La Fortelle, 2017)*

In this kinematic bicycle model, the three DOFs are longitudinal velocity ($V_x$) lateral velocity ($V_y$), and yaw rate ($\dot{\psi}$).

### Dynamic Bicycle Model

Commonly used for modeling with more extreme ARV maneuvers such as an EOAM, is a dynamic bicycle model[35], like that shown by (Attia, Orjuela, & Basset, 2012) (Figure 34). This model includes left and right wheel forces combined into one wheel per axle[36], along with a defined yaw moment, and noted sideslip angle. These combined forces and moments are necessary in developing the dynamic equations of motion for understanding the modeled vehicle responses to various inputs.

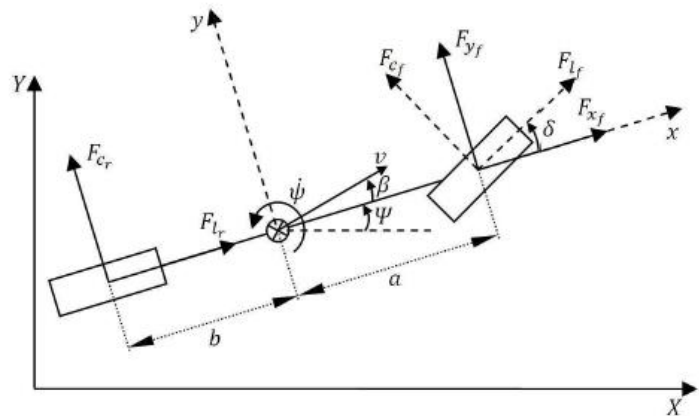

*Figure 34: Dynamic bicycle model representation (Attia, Orjuela, & Basset, 2012)*

A higher DOF "two-track" dynamic model[37] was utilized by (Polack, Altché, d'Andréa-Novel, & de La Fortelle, 2017) for trajectory generation performance comparisons to the simpler kinematic model. This higher DOF two-track model was an adaptation of the dynamic bicycle model defined earlier that includes individual longitudinal and lateral tire forces for each wheel, per vehicle axle (Figure 35).

---

[35] A bicycle vehicle model is often known as a "single-track" model.

[36] Often both longitudinal and lateral tire forces are included at each wheel of even a simplified vehicle model, though some models excluded one or the other, depending on the focus of the modeling application.

[37] A two-track vehicle model is known to be an expansion of the single-track/bicycle model, by instead including four wheels, with at least the front two wheels being capable of steering

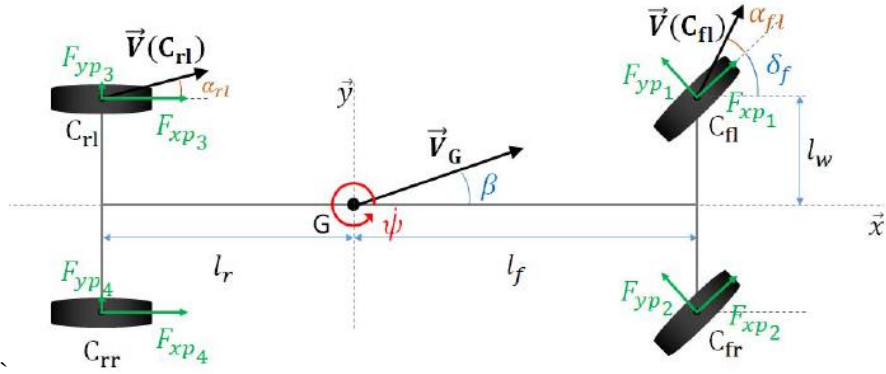

*Figure 35: Two-track dynamic vehicle model (Polack, Altché, d'Andréa-Novel, & de La Fortelle, 2017)*

A table of the associated parameters used for defining the four-wheel/two-track dynamic vehicle model in Figure 35 can be seen in Figure 36.

NOTATIONS

| | |
|---|---|
| $X, Y$ | Position of the center of gravity of the vehicle in the ground framework |
| $\theta, \phi, \psi$ | Roll, pitch and yaw angles of the car body |
| $V_x, V_y$ | Longitudinal and lateral speed of the vehicle in its inertial frame |
| $V_{x_{p_i}}$ | Longitudinal speed of the center of rotation of wheel $i$ expressed in the tire frame |
| $\omega_i$ | Angular velocity of the wheel $i$ |
| $I_x, J_y, I_z$ | Inertia of the vehicle around its roll, pitch and yaw angles |
| $I_{r_i}$ | Inertia of the wheel $i$ |
| $T_{\omega_i}$ | Total torque applied to the wheel $i$ |
| $F_{x_{p_i}}, F_{y_{p_i}}$ | Longitudinal and lateral tire forces expressed in the tire frame |
| $F_{x_i}, F_{y_i}$ | Longitudinal and lateral tire forces expressed in the vehicle frame |
| $F_{z_i}$ | Normal reaction forces on wheel $i$ |
| $M_T$ | Total mass of the vehicle |
| $l_f, l_r$ | Distance between the front (resp. rear) axle and the center of gravity |
| $l_w$ | Half-track of the vehicle |
| $h$ | Heigth of the center of gravity |
| $r_{eff}$ | Effective radius of the wheel |

*Figure 36: Associated parameters for a four-wheel dynamic vehicle model (Polack, Altché, d'Andréa-Novel, & de La Fortelle, 2017)*

While the associated equations of motion (EOMs) and derivations for the Kinematic bicycle (Half) model in Figure 33 and Dynamic Full model in Figure 35, it is left to the reader to review such details. Explanation of these details and more can be found in (Polack, Altché, d'Andréa-Novel, & de La Fortelle, 2017), (Gillespie, 1992), (Genta, 1997), and (Wong, 2008).

### *A Note on Tire Models*

Tire models can be introduced into vehicle model applications, to provide more detailed accounts of the forces generated by the tire-road interactions, for dynamic vehicle models. However, modern tires have complex and detailed construction, in addition to forces and moments generated due to their interactions with the surfaces on which they contact, and the vehicles, to which they are attached (Dixon, 1996), (Pacejka, 2012). These tire forces and moments can be dependent on the vertical load applied from the vehicle, and the type (summer, all-season, snow, off-road), the size, and the rubber compound considered.

Additionally, with respect to properties that affect tire-road friction, both tires and roads can vary with time, and this variance can yield non-uniform performance with respect to time (Gillespie, 1992) (Pacejka, 2012). With that said, tire models must be a compromise between the goals of the output, the accuracy of the model, the complexity of the model, and the computational time required to generate analytical outputs of the relevant tire forces and moments. If a tire model is included in a vehicle model, the type of tire model considered, can have a substantial effect on the computational difficulty of the entire vehicle model, and determine if that vehicle model assumes linear or nonlinear behavior.

In the consideration of high speed, EOAMs, tire models that can generate accurate outputs over a broad range of longitudinal and lateral slip angles, which may exceed the tire's linear behavior range, are necessary. Additionally, tire models for high speed or low tire-surface friction should aim to be computationally light, in the sense that their output forces and moments can be computed in real time[38]. For high speed EOAMs, the need for computational lightness, and accurate representation of linear and nonlinear tire performance are seemingly conflicting constraints. It is this, and similar problems, that have motivated tire modeling solutions that scale, or modify linear tire models to mimic behavior of nonlinear tire models, such as (Talvala & Gerdes, 2008). Other approaches for these tire modeling solutions are to modify nonlinear tire models to be less computationally complex, while maintaining elements of nonlinear tire performance. Such solutions can be reviewed more, in recent tire model survey papers, such as those by (Li, Yang, & Yang, 2014) and (Singh, Arat, & Taheri, 2019).

## Decision-Making Processes for ARVs

After selecting a suitable vehicle model for ARV trajectory generation and/or trajectory-tracking control, during an EOAM, an important next step is the definition of a set of allowable maneuvers for the EOAM. As noted by (Adams & Place, 1994) a set of allowable maneuvers for the EOAM primarily consist of two possible corrective actions: speed control with throttle or brake, and lateral control with steering. Often in cases with human drivers during an EOAM, these corrective actions appear in the form of pure braking[39], pure steering[40], or a combination of braking and steering[41], by the driver (Adams & Place, 1994), (Adams, Flannagan, & Sivak, 1995), (Shiller & Sundar, 1998), (AksunGuvenc, Acarman, & Guvenc, 2003), (Emirler, et al., 2015). While these maneuver options were first proposed for human drivers during an EOAM, they are also suitable for ARVs during EOAMs; they include similar vehicle inputs, applied vehicle systems, and external environment (road and obstacles) involved in the EOAM.

Once all the available options for maneuvers are considered by the Decision-Making subsystem for an ARV during an EOAM, metrics for deciding which maneuver is appropriate must be used in choosing the appropriate maneuver type. The type of maneuver choice for an EOAM can be based on a variety of factors, not limited to: the current ego vehicle states[42], the states of the target obstacle(s), the physical capabilities of the driver or dynamic capabilities of the ARV actuators and controllers, the environmental conditions, traffic volume, and other available knowledge of the scenario for the driver or ARV perception processors. Due to the nature of these many variables considered for the logical decision-making used in the EOAM and other analogous situations, a careful synthesis of the available environmental and vehicle data with the understood vehicle layout and capabilities should be included in the ARV Decision-Making subsystem. Some options for these Decision-Making systems for an ARV during an EOAM are discussed in the subsections below.

### General ARV Decision-Making

While this manuscript has a focus on ARV motion within the context of EOAMs, when reviewing ARV decision-making, it is important to see overall methods that can be used for various types of decisions required for ARVs. This broad understanding of ARV decision-making is important because the way in which the various subsystems function and interact, depend on the system hierarchy applied for the ARV. For example, when an ARV is in a general path planning mode and perceives that an EOAM is necessary, there is a chain of decisions that must occur well before the EOAM would ever be engaged. It is for this reason that we will review general ARV decision-making hierarchies, before moving on to ARV decision-making tailored for EOAMs.

---

[38] The need for real-time calculation or faster, in this sense, is for software-in-the-loop systems (SILS), hardware-in-the-loop systems (HILS), and even for on-vehicle applications.

[39] This includes no other driver inputs besides braking.

[40] This includes either no brakes and no throttle, no brakes and partial (known as maintenance) throttle, or no brakes and no throttle.

[41] This option assumes no throttle is applied.

[42] Some states considered are position, orientation, yaw rate, velocity, acceleration, and jerk.

In the General ARVs section of the Background in this manuscript, a high-level system hierarchy that showed a graphical ARV process flow of logic and data was illustrated in Figure 2. This process flow of data and logic for a general ARV depends on the interaction of four major subsystems:

- Sensing and Perception
- Motion Planning
- Path Generation
- Path-Tracking Control

ARV Decision-Making falls within the Motion Planning system but can also be a component of any subsystem in the global systems discussed in the ARV System Hierarchies section. This can be seen in a general ARV system breakdown shown by (Claussmann, Revilloud, Glaser, & Gruyer, 2017) in Figure 37

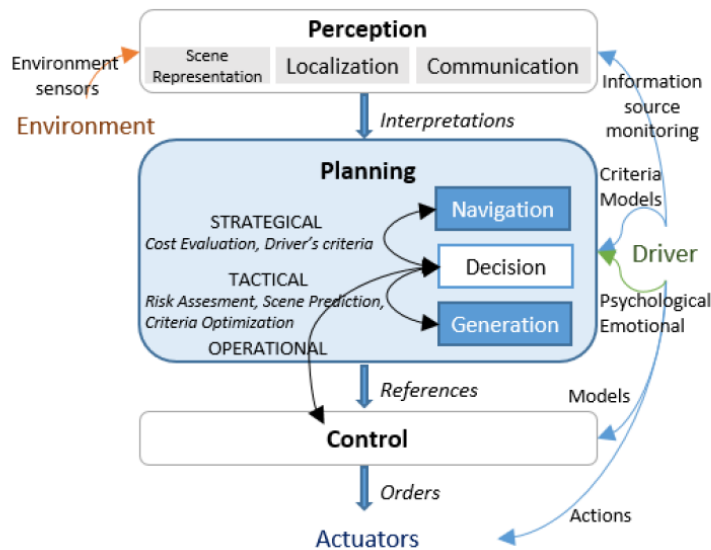

*Figure 37: General ARV System Hierarchy with Decision-Making included in the Motion Planning stage (Claussmann, Revilloud, Glaser, & Gruyer, 2017)*

After a general Motion Planning system uses internal logic to decide that an EOAM is necessary, the EOAM framework of an ARV should utilize a set of discrete steps that it must execute, which are defined in order, below:

- Decision-Making
- Trajectory Generation
- Trajectory-Tracking Control

In the following sections, decision-making approaches which are specifically useful for EOAMs will be reviewed, while trajectory generation and trajectory-tracking control strategies will be discussed.

### ARV Decision-Making Categories

In (Claussmann, Revilloud, Glaser, & Gruyer, 2017), the overall decision-making functions are described as artificial intelligence (AI) for autonomous driving. This introduction to AI in the context of decision-making for ARVs was reviewed on a broader scope in (Claussmann, Revilloud, Gruyer, & Glaser, 2019). The main categories in this overview were Human-like, Heuristic, Approximate reasoning, and Logic, and those categories were distributed on a category plot, organized with a Learning/Rules axis, and Cognitive/Rational axis (Figure 38).

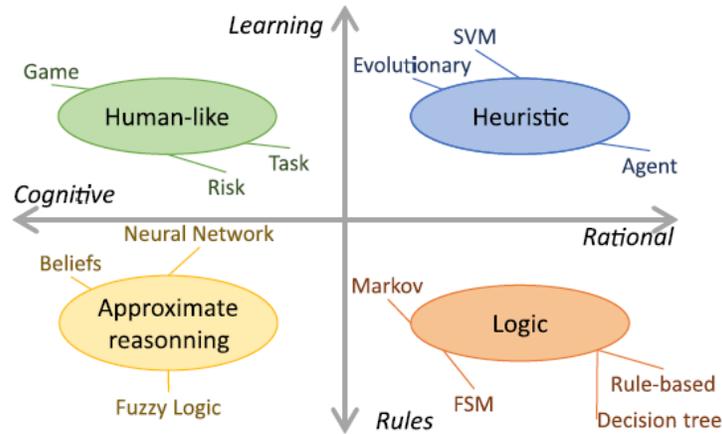

*Figure 38: Category Map of AI categories for ARVs (Claussmann, Revilloud, Gruyer, & Glaser, 2019)*

The broad functions that the AI technologies in Figure 38 offer within the context of ARV decision-making, are to reproduce and simulate human drivers' reasoning and learning. This must be done while the decision-making technologies 'think and act' consistently with the outside environment, in which the ARV interacts (Claussmann, Revilloud, Gruyer, & Glaser, 2019).

Another summary of these decision-making technologies was offered by (Katrakazas, Quddus, Chen, & Deka, 2015), and can be seen in Figure 39[43].

---

[43] The citations provided by (Katrakazas, Quddus, Chen, & Deka, 2015) are the following: (Furda & Vlacic, 2011), (Hardy & Campbell, 2013), (Ziegler et al., 2014), (Kala & Warwick, 2013), (Wei, Snider, Gu, Dolan, & Litkouhi, 2014), (Bandyopadhyay, et al., 2013), (Brechtel, Gindele, & Dillmann, 2014), (Aoude, Luders, Levine, & How, 2010) (Martin, 2013)

| Study | Method | Criteria | Environment description | Drawbacks |
|---|---|---|---|---|
| Furda and Vlacic (2011) | Multiple Criteria Decision Making | Traffic rules Road boundaries Safety distance Collisions Waiting time | Non-intersection segments | Need precise information and manually specified weight for each criterion |
| Hardy and Campbell (2013) | Driving Corridors and Non-Linear Constrained Optimisation | Behaviour towards static and dynamic obstacles Vehicle dynamics Distance to obstacles Distance to goal | Intersections | Computational effort rises with number of obstacles <br><br> Ignorance of social interactions between traffic participants |
| Ziegler et al. (2014b) | Driving Corridors and Hierarchical State Machines | Static and dynamic obstacles behaviour Yield and Merge rules | Intersections and non-intersection segments | Other vehicles are presumed not to accelerate and to keep safe distances from road boundaries |
| Kala and Warwick (2013) | Behaviour Choice according to Obstacle Motion | Distance and velocity constraints | No road lanes | Infinite straight roads Problems on curvy roads, overtaking and conflicting behaviours |
| Wei et al. (2014) | Prediction and Cost-function | Comfort, safety, fuel consumption, distance to goal | Straight roads | Only single-lane behaviours tested |
| Bandyopadhyay et al. (2012) | Mixed-Observability MDP | Pedestrian position and velocity | Pedestrian crossings | Different modelling required for each pedestrian Intentions are assumed unchangeable |
| Brechtel et al. (2014) | Partially Observable MDP | Vehicle position and velocity | Merging scenarios with occluded vision | Continuous belief space may lead to large number of samples and large computational effort |
| Aoude et al. (2010a) | Game Theory | Time to collision | Intersections | Model-car evaluation at low speeds Perfect information assumed |
| Martin (2013) | Game Theory | Position, speed, acceleration and manoeuvre choice | Straight roads | Perfect information required |

*Figure 39: A summary of various planning approaches with emphasis on decision-making and obstacle prediction (Katrakazas, Quddus, Chen, & Deka, 2015)*

While the decision-making and obstacle prediction methods summarized by both (Katrakazas, Quddus, Chen, & Deka, 2015) and (Claussmann, Revilloud, Gruyer, & Glaser, 2019) are quite relevant to ARV general path planning, the methods within the Logic category from Figure 38 are most relevant to EOAMs. The reasons why Logic-based decision-making strategies are best suited for EOAMs, especially EOAMs at higher speeds or lower levels of available tire/road friction, are reviewed in the EOAM Decision-Making Logic section, below.

Considering the decision-making methods shown in Figure 38 and Figure 39 which fall outside of the Logic category, while still important for the development of ARV general motion planning, they are outside of the scope of this manuscript. However, for a brief summary, some of these decision-making methods outside of the Logic category from Figure 38 are: driving corridors for planning in urban settings (Ziegler et al., 2014), highway lane change using adaptive analytic hierarchy process (AHP) and fuzzy logic (Kim & Langari, 2012) or using Markov decision process (MDP) and game theory (Aoude, Luders, Levine, & How, 2010), (Bandyopadhyay, et al., 2013), (Martin, 2013), (Brechtel, Gindele, & Dillmann, 2014), (Coskun S. , 2018). Some other methods are the Gaussian modeling and reinforcement learning (Sadigh, Sastry, Seshia, & Dragan, 2016), as well as Heuristic Agents (Schwarting, Alonso-Mora, & Rus, 2018).

*Object and Obstacle Motion Prediction*

When considering environmental object and obstacle risk-prediction (Lefèvre, Laugier, & Ibañez-Guzmán, 2012) (Lefèvre, Vasquez, & Laugier, 2014) reviewed decision-making methods such as: Heuristics, Relevance Vector Machines (RVM), Support Vector Machines (SVM), Multi-Layer Perceptrons (MLP), Logistic Regression, Hidden Markov Model (HMM), and more. Terminology for object collision risk assessment was also provided, such as time-to-collision (TTC), time-to-react (TTR), time-to-intercept (TTI), time-to-zone of clearance (TTZ), or general, time-to-x (TTX) (Lefèvre, Vasquez, & Laugier, 2014), (Claussmann, Carvalho, & Schildbach, 2015). The above references should be considered for exploration of decision-making within ARV motion planning outside of consideration of an EOAM at higher ego vehicle speeds or lower tire/road friction.

## EOAM Decision-Making Logic

Within the Logic portion of the ARV AI Map for decision-making provided by (Claussmann, Revilloud, Gruyer, & Glaser, 2019)(Figure 38), the sub-categories for decision-making that are most relevant for EOAMs are Finite State Machines (FSMs), Hierarchical State Machines (HSM), Rule-Based (methods), and Decision Tree. The FSM/HSM[44], Rule-Based, and Decision Tree sub-categories are commonly used in EOAM-related decision-making because of their discrete sets of available options/combinations. This discrete nature allows these FSM/HSM, Rule-Based, and Decision Tree sub-categories to be computationally light (Furda & Vlacic, 2011) compared to the other decision-making sub-categories from Figure 38 and Figure 39; in some cases, computations from these sub-categories can be completed off-line, such that they serve as a look-up table for online computations. Overall, the discrete nature, computational lightness, and availability of some off-line computations, lends the Logic-based sub-categories of ARV decision-making (Claussmann, Revilloud, Gruyer, & Glaser, 2019) to be useful in the highly dynamic and time-critical EOAMs.

The use of an FSM or HSM is advantageous for time-critical decision-making systems—like those designated for EOAMs—due their logic which consists of initiating and operating multiple state in series or in parallel, while some include the use of "macros to generate complex actions" (Argall, Browning, & Veloso, 2007). The FSMs are noted to be different from the Rule-Based decision-making due to the use of a "predetermined sequence of actions and states" (Claussmann, Revilloud, Gruyer, & Glaser, 2019). One example of this, which involves the identification of discrete maneuver states which are navigated by predefined maneuver switching rules integrated into path generation and control for simulated highway driving, can be seen in Figure 40, where $C_i$, $E_i$, $F_i$ are denoted as specific switching conditions used to transition between the various vehicle states $S_i$ (Wang, Weiskircher, & Ayalew, 2015).

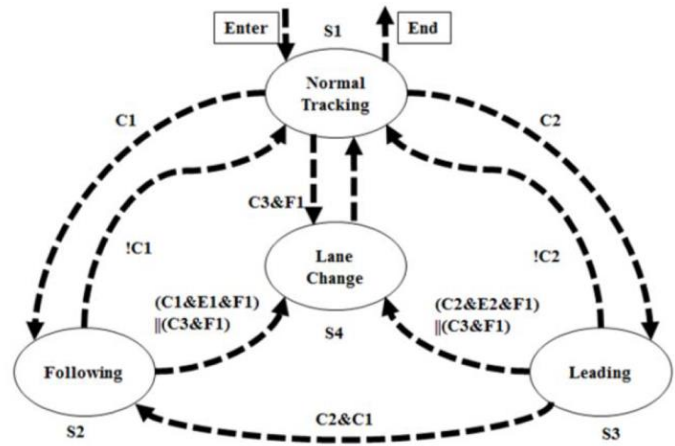

*Figure 40: Example of a Finite State Machine demonstrated by (Wang, Weiskircher, & Ayalew, 2015), with Ci, Ei, Fi denoted as specific conditions to transit between the vehicle behaviors Si*

It is noted that authors such as (Argall, Browning, & Veloso, 2007) make the distinction between FSMs and HSM, by denoting an HSM as an FSM organized with defined hierarchical levels. One example of an HSM with two defined levels for lane changes during highway driving is shown by (Xiong, et al., 2018), where the high-level states and switching rules are shown in Figure 41 and the corresponding substates are shown in Figure 42.

---

[44] In many articles, the finite state machine is also referred to as the hierarchical state machine (HSM), though some differentiate HSMs as being FSMs designated into specifically defined tiers (Argall, Browning, & Veloso, 2007)

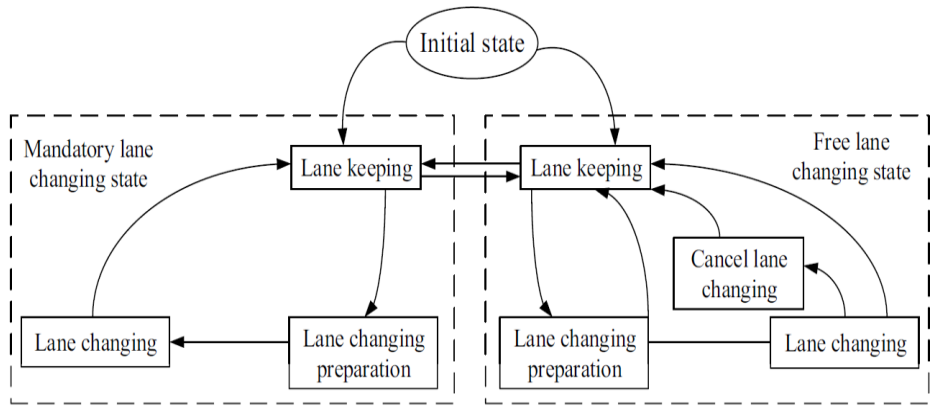

*Figure 41: High-level states for highway-driving lane changes of an ARV (Xiong, et al., 2018)*

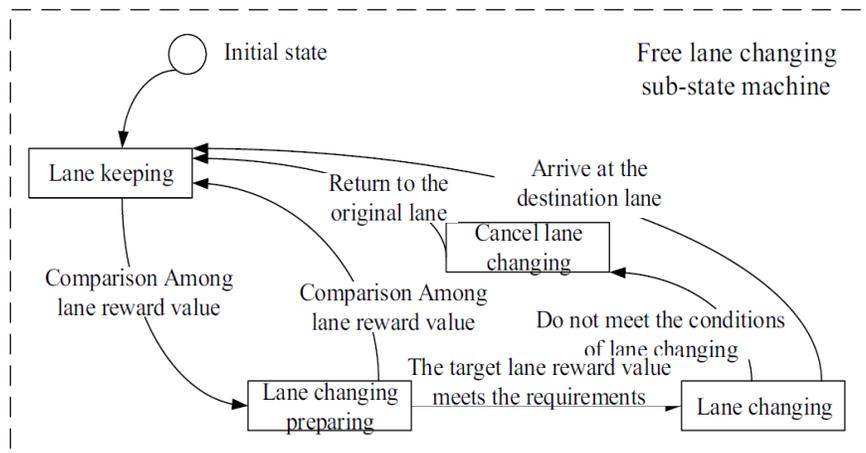

*Figure 42: HSM substates highway-driving lane changes of and ARV (Xiong, et al., 2018)*

Looking at a Decision Tree decision-making method, (Claussmann, Carvalho, & Schildbach, 2015) utilized high-level and low level decisions, with logical questions at each level, that could be answered with a "yes" or "no" response, based on various sensor data, environmental perception, and vehicle states (Figure 43).

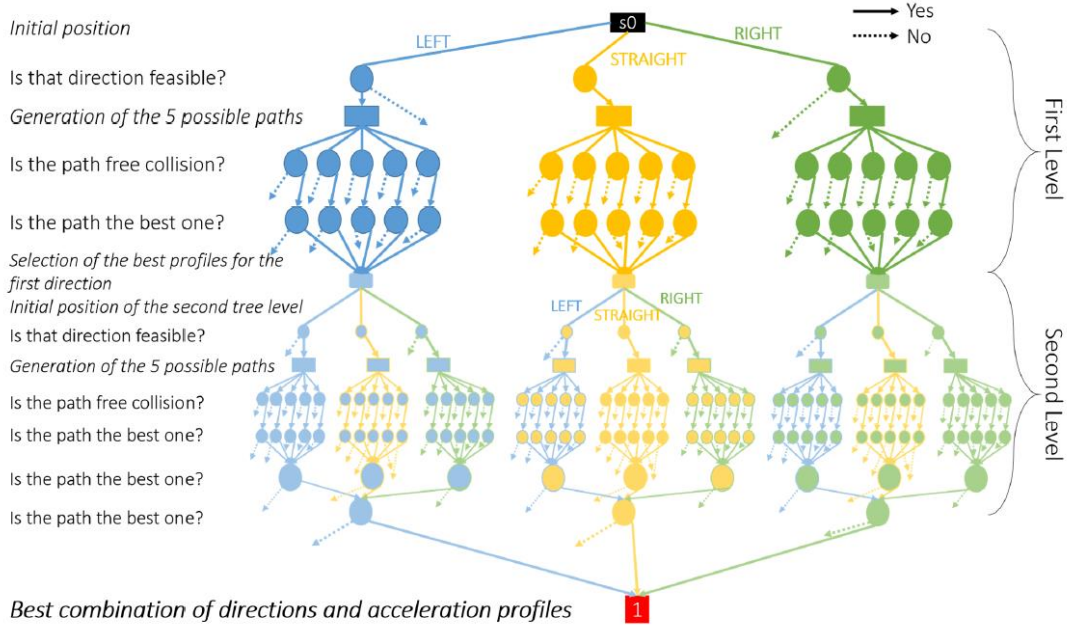

*Figure 43: Example of a Binary Decision Diagram (BDD) used for highway lane-change decison-making (Claussmann, Carvalho, & Schildbach, 2015)*

One application of the Rule-Based sub-categories of ARV decision-making, is the Multiple Criteria Decision-Making (MCDM) method utilized by (Furda & Vlacic, 2011) as also shown by (Katrakazas, Quddus, Chen, & Deka, 2015), in Figure 39. Looking deeper into the method applied by (Furda & Vlacic, 2011), Figure 12 shows a detailed breakdown of the high level ARV system architecture introduced in Figure 12.

This decision-making and overall software architecture flow diagram show not only how the Real Time (RT) Decision-Making and Driving Maneuver Control subsystem integrates into the broader ARV system, but also indicates the various feedback signals that supply inputs and offer outputs for all the mid-level subsystems. Within the context of an EOAM, (Furda & Vlacic, 2011) describes two discrete sub-decisions that should be made within the RT Decision Making subsystem:

1) **Safety Check**: discovering a set of feasible maneuvers or target set of behaviors (Claussmann, Revilloud, Glaser, & Gruyer, 2017) that provide an opportunity to avoid the perceived obstacle, while obeying local traffic laws, and adhering to environmental constraints, such as other vehicles, VRUs, curbs, barriers, weather, and more

2) **Optimal Maneuver Selection**: from the subset of choices offered from 1), decided the optimal maneuver based on the available driver inputs to the vehicle (throttle, steering, braking)

Within the context of an EOAM, both the Safety Check and Optimal Maneuver Selection are executed in sequential order, and considered regardless of whether an FSM/HSM, Decision Tree, or Rule-Based decision-making method is utilized. However, determining which maneuvers are safe and feasible, then selecting the optimal maneuver from that set is very much in line with how a human driver performs an EOAM. This human driving decision-making process is often used to aid the development of the details for a decision-making algorithm for ARVs. The relationship between human-based EOAM decision-making and more detailed criteria for EOAM decision making will be discussed in the Deciding Feasible Maneuvers for an EOAM section, below.

## Deciding Feasible Maneuvers for an EOAM

### Human-Based EOAM Studies for Decision-Making: Review

As mentioned earlier, the same fundamental driver input options and external environment considerations during an EOAM for an ARV are relevant for a human driver. An early review of studies involved with measuring or simulating human driver decision-making during a EOAM was conducted by (Adams & Place, 1994). In that review, experiments from both field tests and driving simulators were surveyed, as well as studies that analyzed accident-avoidance strategies from actual accident data.

The primary conclusions gained from both the human EOAM test studies and the accident data review studies were the following:

- Drivers are more likely to brake than steer for an EOAM, though providing steering only or steering with braking, is more optimal to avoid the obstacle
- Either steering only, or steering and braking, are better than only braking, during an EOAM
- Drivers have reluctance to steer during EOAM situations[45]
- The tendency to provide steering as an input is higher, at higher vehicle speeds and lower relative distances between the ego vehicle and approaching the obstacle

With these overall conclusions (Adams & Place, 1994) mentioned that based on these results of the review, there are opportunities for vehicles that can do this on their own, implying that ARVs could use the conclusions to create their own EOAM strategies.

*Human-Based EOAM Studies for Decision-Making: Simulator Study*

Later, (Adams, Flannagan, & Sivak, 1995) went on to make their own driving simulator study for an EOAM, where the goal was to determine whether drivers are more likely to steer, brake or apply both as inputs during an EOAM scenario. This scenario included participants driving on a rural two-lane road, where their task was to avoid a rock that appeared in road while they were traveling at a speed of 89 kph, before the obstacle appeared, at a preview distance of 49 m; trials were both alerted and un-alerted.

Similar results to the literature survey provided by (Adams & Place, 1994) were expected to be true with live participants in the simulator study which followed (Adams, Flannagan, & Sivak, 1995). However, (Adams, Flannagan, & Sivak, 1995) found that when attempting to avoid the obstacle during unalerted trials, drivers had more success when braking and steering, followed by steering only, and no success with braking only. Additionally, it was found that the most popular choice for participants, was to steer only. It was noted by (Adams, Flannagan, & Sivak, 1995) that these results which showed pure steering to be more popular to steering and braking, or pure braking, were likely due to the sudden obstacle (modeled as a rock) being small relative to the lane size, the lack of realism in the driving simulator, and that there was no oncoming traffic in the adjacent lane. The authors still concluded that regardless of the maneuver performed during an EOAM, there would be several advantages to allowing an ARV to execute these maneuvers that a driver either is not capable of executing or does not want to execute, even though they are the correct method to avoid certain obstacles (Adams, Flannagan, & Sivak, 1995).

*Human-Based EOAM Studies for Decision-Making: Modeling*

One early decision-making model simulation which echoed the options for an EOAM mentioned by (Adams & Place, 1994), was defined in a decision-making flow diagram for an ARV, called the driver-vehicle effectiveness model (DRIVEM) (Wolf & Barrett, 1978). This decision-making flow was based on human logic and designed to first a) detect a critical event, b) decide if action is required, c) apply open or closed-loop control to perform the action, and d) return back to original autonomous driving control, once the desired final trajectory has been achieved (Wolf & Barrett, 1978), (Reid L. , 1983). With the DRIVEM, if part b) deemed that action was required, the maneuvers considered would be the same as those listed by (Adams & Place, 1994), which were pure steering, pure braking, or a combination of steering and braking. The decision for which maneuver to perform would be based on the external conditions and would determine the braking ($\delta_b$) and/or steering inputs ($\delta_s$), which would look like the time histories in Figure 44.

---

[45] This was noted to be caused by drivers wanting to maintain their lane in general, and drivers not understanding how to perform a steering avoidance maneuver while not fully understanding their vehicle's handling capabilities (Adams & Place, 1994)

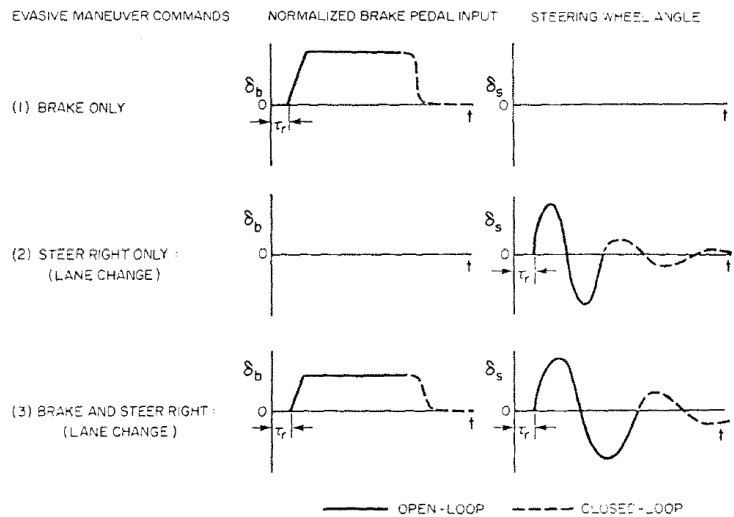

*Figure 44: Sample time histories for various EOAMs considered in the DRIVEM (Wolf & Barrett, 1978), (Reid L. , 1983)*

Thus, a clearly determined set of feasible input modes for an ARV during an EOAM, derived from human-based logic, was presented by (Wolf & Barrett, 1978), (Adams & Place, 1994), and (Adams, Flannagan, & Sivak, 1995) to be: steering, braking, or combined steering and braking. As a next step, determination of the conditions under which steering, braking, or combined steering and braking are appropriate, must be found. Methods that are well-suited for determining the suitable conditions that determine the input mode(s) for an EOAM are discussed more in the Detailed Criteria for EOAM Feasible Maneuver Options section.

## Detailed Criteria for EOAM Feasible Maneuver Options

As mentioned in the Deciding Feasible Maneuvers for an EOAM section, when considering an EOAM for an ARV, the three available modes of actuator inputs that are useful for the ego vehicle to avoid obstacle(s) within the region of interest (ROI)[46] are: steering, braking, or combined steering and braking. After acknowledging these set of inputs for an EOAM, the next step is to understand when, and under what circumstances, each input is most appropriate.

### Defining Ranges for Appropriate EOAM Inputs

One study that attempted to answer this question was constructed by (Shiller & Sundar, 1998), which introduced the idea of Clearance Curves. The Clearance Curves were defined in a space comparing relative distance to an active roadway obstacle[47] (ARO) versus ego ARV longitudinal speed in a phase portrait, for which steering, braking, or combined steering and braking would be appropriate driving inputs (Figure 45).

---

[46] In the case of an EOAM, the ROI is the area for forward viewing road in which an obstacle is present, or on a trajectory to enter, such that the ego ARV must take corrective action to avoid the obstacle

[47] In this sense an ARO could be another vehicle, VRU, or significant inanimate object that poses a danger to the ARV, due to entering the roadway

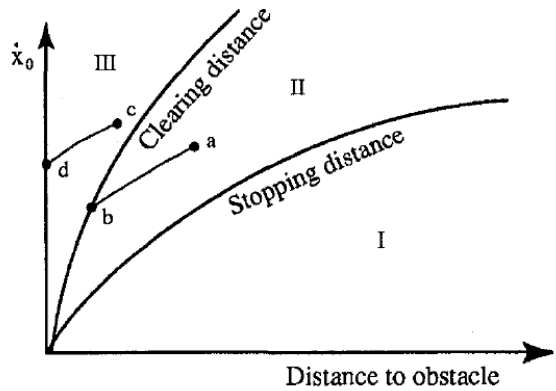

*Figure 45: Clearance curves in the distance-to-obstacle versus longitudinal speed, phase-plane (Shiller & Sundar, 1998)*

Defined by (Shiller & Sundar, 1998) in Figure 45, Region *I* represents the phase of the relative ARO distance versus ego ARV longitudinal speed, when the ego ARV has sufficient time[48] to come to a complete stop with braking only or a "relaxed lane transition." When an object is detected by the ego ARV in Region *II*, a lane change with steering, or combined steering and braking are the available options to avoid a collision. The transition from point *a* to point *b* in the phase-plane represents the execution of braking first to lower vehicle speed before performing a lane change at the Clearance Curve boundary condition. This transition would allow for the ego ARV to have more time in the current lane, if environmental (traffic) conditions suggested this was favorable.

Region *III* denotes the phase-plane conditions under which a head-on collision with the obstacle in the ROI is imminent, and the best actuator input is pure braking, to lessen the severity of the head-on collision (Shiller & Sundar, 1998). It is suggested by (Shiller & Sundar, 1998) that attempting a lane change while the ego ARV is in Region *III* would result in an off-center collision with the object in the ROI, or general loss of control during the lane change, both of which are "generally more dangerous than a head-on collision, for which the vehicle is better designed to sustain." While Figure 45 exhibited a general Clearance Curve phase portrait, (Shiller & Sundar, 1998) offered a phase portrait based on a dynamic point-mass vehicle model in which maneuvers were determined using optimal control methods.

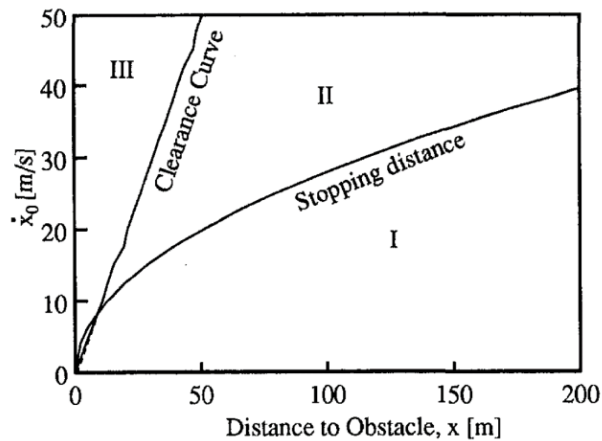

*Figure 46: Clearance Curve phase portrait utilizing a dynamic point-mass model (Shiller & Sundar, 1998)*

It is notable that below an ARV speed of nearly 10 m/s, in Figure 46 the choice of either pure steering or pure braking as driving inputs would result in an avoided impact, depending on the relative distance to the ARO.

---

[48] The use of time here can be considered equivalent to time-to-collision (TTC)

It is also noted that, within the phase portrait regions of suitable driving inputs for an EOAM, the Clearance Curves also indicate the usable distance detection ranges for ARV on-board distance sensors, as well as the performance of the ARV in braking and handling (Shiller & Sundar, 1998). A shift in performance of the sensing and perception capabilities, as well as the braking and handling performance of the ARV, would subsequently shift and modify the shapes of the Clearance Curves.

*Defining Minimum Distances for EOAM Actions*

In addition to defining transition points for various suggested driving inputs during the decision-making process of an EOAM, (Shiller & Sundar, 1998) discussed the idea of the minimum clearance distance, which is the distance at which a point on the ego ARV intersects with a point on the ARO. Others have adopted this same concept of a minimum distance needed to execute an EOAM without impacting an ARO. A Minimum Avoidable Distance for an EOAM was proposed by (Hattori, Ono, & Hosoe, 2006); however, this minimum avoidable distance introduced was characterized for pure braking or steering driving inputs, only. Still, this Minimum Avoidable Distance concept was similar to that of (Shiller & Sundar, 1998) in that there existed definable points in distance and time to execute one of the various driving inputs of an EOAM, beyond which there would be an impact between the ego ARV and the ARO. This was demonstrated by Hattori et al. 2006 with distance $X_{es}$ equating the minimum stopping distance to avoid an impact with the ARO (Figure 47 – a), and distance $X_{ep}$ equating the minimum distance to perform a passing (steering) EOAM to avoid an impact with the ARO (Figure 47 – b),

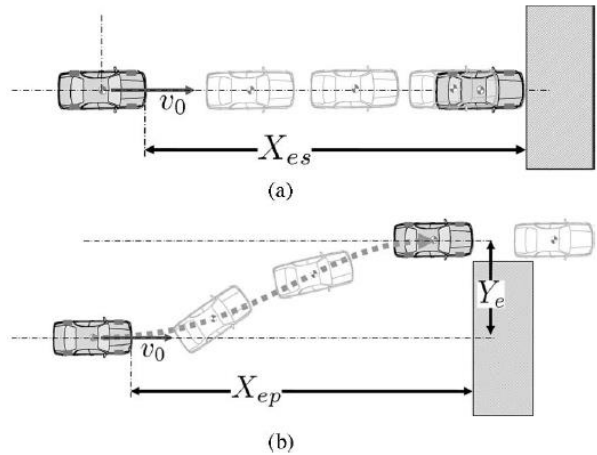

(a)

(b)

*Figure 47: Minimum stopping ($X_{es}$ – a) and passing ($X_{ep}$ – b) distances for an ARV EOAM to avoid an impact with an ARO (Hattori, Ono, & Hosoe, 2006)*

Like (Shiller & Sundar, 1998), (Hattori, Ono, & Hosoe, 2006) also constructed a phase diagram with the Minimum avoidable distance and initial longitudinal speed of the ARV during an EOAM (Figure 48).

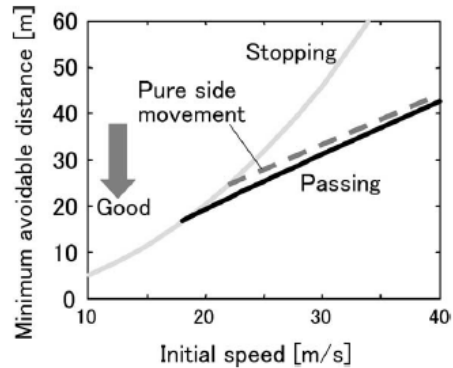

*Figure 48: Minimum Avoidable Distance vs. Initial Speed phase diagram for an EOAM (Hattori, Ono, & Hosoe, 2006)*

The axes in the phase diagram by Hattori et al., 2006 are switched compared to that by (Shiller & Sundar, 1998), but the outcomes are objectively similar. Also, similar to (Shiller & Sundar, 1998), (Hattori, Ono, & Hosoe, 2006) noted that at up to a certain ARV speed (18.6 m/s) either pure steering or pure braking would be acceptable driving inputs to avoid impact during the EOAM. The choice of driving input would be determined based on the Minimum Avoidable Distance to the ARO. Hattori et al., 2006 do not discuss the possibility of the combination of steering and braking combined for driving inputs during an EOAM, though this intermediate choice can be seen after speeds of 18.6 m/s.

In a recent study by (Peng, et al., 2020) a Minimum Safety Distance was utilized as a key part during the decision-making process of an EOAM. This Minimum Safety Distance was analogous to the Minimum Avoidable Distance (Hattori, Ono, & Hosoe, 2006) (Wang, Cui, & Yu, 2019), and the relative distance in the Clearance Curve phase portrait (Shiller & Sundar, 1998). One additional aspect in the study by (Peng, et al., 2020) was the utilization of the Minimum Safety Distance for an EOAM decision-making process that involved scoring to determine the chosen maneuver.

### Utilizing Scoring in Decision-Making

In the EOAM decision-making process with scoring that was demonstrated by (Peng, et al., 2020), either steering, or a combined steering and braking maneuver was modeled; pure braking was not included as an available option in the decision-making process. However, (Peng, et al., 2020) include a metric for 1) safety – Minimum Safety Distance for 2) comfort – steering frequency, and maximum lateral acceleration. The safety metric, based on Minimum Safety Distance, is the first metric considered, as safety determines the EOAM feasibility, and is given a higher priority over comfort. This safety metric determines whether the maneuver should include steering with constant longitudinal acceleration ($a_x = 0$), or non-constant one (i.e., with braking) ($a_x \neq 0$).

In this decision-making methodology by (Peng, et al., 2020), there is an offline scoring look-up table used to determine if 1) the maneuver is feasible due to safety decision-making rules and b) how comfortable each maneuver is, that is deemed to be feasible from part 2). An example set of EOAM decision-making scores are shown based on a scenario where the relative speed between the ego ARV and the longitudinally traveling ARO vehicle in Figure 49. Given this relative speed, a mix of steering maneuvers, with and without braking included, are given, based on the steering input frequency. From this steering input frequency, the maximum lateral acceleration during the maneuver can be calculated ($a_{Yc}$). After $a_{Yc}$ is calculated, the difference between relative distance[49] and the minimum safety distance with braking ($L_{m2}$) or without braking ($L_{m1}$) is calculated as $\Delta L$. The total score is $s_1 = s_2 * s_3$, where $s_2$ is either equal to 1 if $\Delta L \geq 0$ or 0 if $\Delta L < 0$. If $s_2 = 0$ the maneuver is not used, but if $s_2 = 1$, the then best maneuver is chosen based on comfort ($s_3$). The comfort for feasible maneuvers is based on the maximum lateral acceleration ($a_{Yc}$), where a smaller magnitude of $a_{Yc}$ achieves a higher score, and $a_{Yc}$ is directly affected by the steering frequency used during the maneuver. Once the highest score of all the maneuver options based on feasibility, and comfort is found ($s_3$), this is selected for the drive mode during the EOAM.

---

[49] In this instance, and all instances moving forward, relative distance is the distance between the ego ARV and the ARO

| Hierarchy | Performance | Driving mode steering frequency Hz/brake deceleration (m/s²) | Maximum acceleration $a_{Yc}(t)_{max}/(m/s^2)$ | Minimum safety distance $L_m$/m | Evaluation parameter | Score |
|---|---|---|---|---|---|---|
| 1 | Overall | – | – | – | – | $s_1 = s_2 \times s_3$ |
| 2 | Safety | – | $\leq a_{Ymax}$ | $\Delta L \geq L_m$ | Difference between relative distance $\Delta L$ and minimum safety distance $L_m$ | $s_2 = 1 \ (\Delta \geq 0)$ |
| | | | | | | $s_2 = 0 \ (\Delta < 0)$ |
| 3 | Comfort | 0.1/0 | 0.2443 | 178.1840 | Maximum acceleration | $s_3 = 10$ |
| | | 0.2/0 | 0.8753 | 103.4900 | | $s_3 = 9$ |
| | | 0.3/0 | 1.7742 | 78.5870 | | $s_3 = 8$ |
| | | 0.4/0 | 2.8562 | 66.1550 | | $s_3 = 7$ |
| | | 0.1/4 | 4.0057 | 130.8748 | | $s_3 = 6$ |
| | | 0.5/0 | 4.0604 | 58.6940 | | $s_3 = 5$ |
| | | 0.2/4 | 4.0807 | 89.8758 | | $s_3 = 4$ |
| | | 0.3/4 | 4.4389 | 68.1770 | | $s_3 = 3$ |
| | | 0.4/4 | 5.0678 | 60.0598 | | $s_3 = 2$ |
| | | 0.5/4 | 5.8642 | 53.3588 | | $s_3 = 1$ |

*Figure 49: Decision-making lookup table of various EOAM variants, and their corresponding scores (Peng, et al., 2020)*

Though not formally mentioned by (Peng, et al., 2020), this type of rule-based (and scoring-based) ARV decision-making strategy multiplied a series of output score values based on discrete criteria, then summed them to get a final score that falls within the category of Multiple Criteria Decision Analysis (MCDA), or Multiple Criteria Decision-Making (MCDM) (Churchman, Ackoff, & Arnoff, 1957) (Yoon & Hwang, 1995). One similar aspect of MCDM that was also used in decision-making for an ARV is called Simple Additive Weighting (SAW). This SAW method was employed by (Furda & Vlacic, 2011) as mentioned in the EOAM Decision-Making Logic section.

The SAW method was utilized by (Furda & Vlacic, 2011) for the 2) Optimal Maneuver Selection stage, once the list of feasible maneuvers were selected in the 1) Safety check stage of (EOAM Decision-Making Logic section). To leverage MCDM with SAW the following features must be defined:

**Objectives** – most general high-level hierarchy goals for an EOAM, with associated sub-level goals (e.g., Avoid AOR in road)

**Attributes** – lower-level actionable breakdowns of the objective

**Alternatives** – different methods by which the objectives can be executed; i.e., different approaches to the same goal (the set of alternatives cover all the possibilities of completing the feasible maneuvers determined from the Safety Check decision-making stage)

**Utility Function** – a relative rating of all the available attributes, with respect to the how well they will fulfill the lowest level of driving objective, which are each of the alternatives.

**Attribute Weights** – multiplicative values associated with each attribute based on its relative importance, given discrete environmental conditions provided by the world model (e.g., the weights would be different based on the scenarios of: a) a large lane width, more than two lanes, sunny, warm, no oncoming traffic, and b) a small lane width, with two lanes, cold, slippery, and no oncoming traffic (still)).

**Decision Matrix** – is a way of defining the evaluation value of each alternative, as the output from the utility function (i.e., each alternative is evaluate with respect to the same attributes) (Yoon & Hwang, 1995)

Below is an example from (Furda & Vlacic, 2011) of a Decision Matrix that includes the culmination of determined Objectives, associate Attributes, various Alternatives, a Utility Function evaluation of the various Attributes, Attribute weights, and the total sum for each alternative, using SAW.

| $a_i$ | $attr_1$ | $attr_2$ | $attr_3$ | $attr_4$ | $attr_5$ | $attr_6$ | $attr_7$ | $attr_8$ | $attr_9$ | $attr_{10}$ | $attr_{11}$ | $V(a_i)$ |
|---|---|---|---|---|---|---|---|---|---|---|---|---|
| $a_1$ | 1 | 0.5 | 0.5 | 0.5 | 0.25 | 0.25 | 1 | 0.5 | 0.75 | 0.75 | 1 | 11 |
| $a_2$ | 1 | 0.25 | 0.5 | 0.5 | 1 | 1 | 1 | 0.5 | 0.75 | 0.75 | 1 | 12.25 |
| $a_3$ | 1 | 0.5 | 0.5 | 0.5 | 0.25 | 0.25 | 1 | 1 | 0.25 | 1 | 1 | 12 |
| $a_4$ | 1 | 0.25 | 0.5 | 0.5 | 1 | 1 | 1 | 1 | 0.25 | 1 | 1 | **13.25** |
| $a_5$ | 0.5 | 0.5 | 0.25 | 0.5 | 0.25 | 0.25 | 0 | 0 | 1 | 0 | 0 | 4.5 |
| $a_6$ | 0.5 | 0.5 | 1 | 0.5 | 0.75 | 1 | 1 | 0.25 | 1 | 0 | 0 | 8 |
| Weight | $w_1 = 1$ | $w_2 = 1$ | $w_3 = 2$ | $w_4 = 1$ | $w_5 = 1$ | $w_6 = 1$ | $w_7 = 1$ | $w_8 = 3$ | $w_9 = 2$ | $w_{10} = 2$ | $w_{11} = 2$ | |

*Figure 50:Simple Additive Weighting example with attributes, alternatives, and summed utility output (Furda & Vlacic, 2011)*

Like the scoring map created by (Peng, et al., 2020), this Decision Matrix by (Furda & Vlacic, 2011) can be computed offline, for various environments which are perceived by the ARV, as well as situational characteristics that would be useful for a given Weighting Map. The one considerable drawback from using the SAW method with a Decision Matrix within MCDM (and rule-based decision-making algorithms in general), is that the various features listed above must be discretized by a human, and can require many cause-and-effect rules to cover the many scenarios which an ARV might see during an EOAM (Claussmann, Revilloud, Gruyer, & Glaser, 2019). However, when MCDM is specifically applied to an EOAM, there can be fewer required environmental variables to consider, compared to general path planning. In future studies, a combination of MCDM for EOAM decision-making with an automated method (such as neural networks or reinforcement learning (Schwarting, Alonso-Mora, & Rus, 2018)) for estimating potential environmental scenarios and MCDM features, could be a useful solution to the current limitation of requiring human-picked rules, in rule-based decision-making methods like MCDM.

*ARO Motion Effects on Decision-making*

While the state-of-the-art of ARO motion prediction algorithms for EOAMs will not be reviewed in this manuscript, one important note regrading ARO motion is that many EOAM-related studies either include avoidance strategies for a longitudinally moving ARO, or a laterally moving ARO, but not both. Studies like (Maeda, Irie, Hidaka, & Nishimura, 1977), (Reid, Graf, & Billing, 1980) have AROs that move laterally entering the roadway in front of the ARV or driver. Meanwhile, others only have obstacles (such as traffic vehicles) that suddenly appear or stop in front of the driver (Shiller & Sundar, 1998), (Adams, Flannagan, & Sivak, 1995), (Hattori, Ono, & Hosoe, 2006), (Peng, et al., 2020), which means the ARO is only moving in the longitudinal direction with respect to the driver/ego ARV. Including both longitudinally and laterally moving ARO is important to fully characterize the capabilities of the obstacle avoidance maneuver design and control.

In a study by (Soudbakhsh, Eskandarian, & Moreau, 2011), there were two different sets of driving scenarios given to the participants during the study, called "A" and "B." Each of the two scenarios "A" and "B" had three near collision situations, which were: 1) the appearance of a car in front of the driver, 2) crossing pedestrians stopped in the right lane, and 3) a parked car pulling over in front of the subject vehicle.

The other scenario ("B") has a car appearing in the road, a truck, which comes from the right side of the road and stops halfway (occupy the whole lane), and the third one is a parked car pulling over. The difference between A and B is for the second step, pedestrians, vs a truck, entering the lane occupied by the driver. The results regarding which option of maneuver the participant driver chose, based on the type of obstacle (and its direction of travel) that entered the road[50] can be seen in Figure 51.

---

[50] The pedestrian object was not reported in this chart

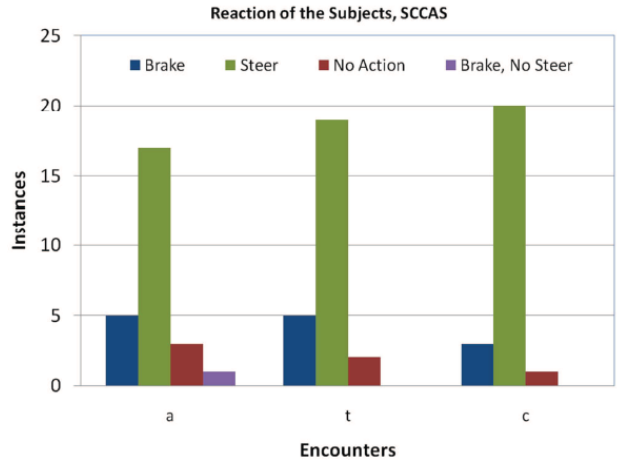

*Figure 51: Participant evasive reaction to obstacles in the roadway, with types of obstacles – a: appearing vehicle, t: crossing truck, and c: car pulling over (Soudbakhsh, Eskandarian, & Moreau, 2011)*

Based on these considerations by (Soudbakhsh, Eskandarian, & Moreau, 2011), future works addressing EOAM development should consider various modes of AOR motion into the path of the ARV, during a scenario which requires an EOAM to avoid a collision.

Considering the sections above regarding decision-making for an EOAM, the necessary steps include the 1) Safety Check, and 2) Optimal Maneuver Selection, involving the available driving inputs of steering, braking, or combined steering and braking. Once the suitable EOAM driving inputs have been selected from a deliberate decision-making process, the next step is to generate a proper EOAM trajectory for the ARV to follow. This will be discussed in the next section, Trajectory Generation for Emergency Maneuvers.

## Trajectory Generation for Emergency Maneuvers

Proper generation of a path for an ARV during general path planning has been studied and reviewed for many years and was covered in the General ARV Path Planning section above. A review of various trajectory generation methods for ARV EOAMs will be discussed next.

### Differences Between Path and Trajectory Generation

Important differences exist between ARV path generation and trajectory generation methods as well as their overall purposes. General path generation techniques can often ignore dynamic effects and utilize geometric or kinematic vehicle modeling, for vehicle motion at lower speeds and lateral dynamics. Some of the common general path generation techniques include planning methods such as Variational Methods, Graph Search-Based and Sampling-Based Planning, for a rough (often discontinuous) navigation through the configuration space (Paden, Čáp, Yong, Yershov, & Frazzoli, 2016). After the rough path has been generated, path smoothing techniques are subsequently utilized to output the ARV's reference path to the actuator controller. These general path planning methods are suitable at lower speeds but do not fit the kinodynamic criteria needed for an EOAM at higher vehicle speeds (Katrakazas, Quddus, Chen, & Deka, 2015), (Claussmann, Revilloud, Gruyer, & Glaser, 2019), (González, Pérez, Milanés, & Nashashibi, 2015), (Sharma, Sahoo, & Puhan, 2019.

The generation of an ARV trajectory differs from the generation of an ARV path, in that trajectories are specifically designed for dynamic maneuvers, in dynamic environments, where vehicle inertial effects cannot be ignored (Paden, Čáp, Yong, Yershov, & Frazzoli, 2016), and dynamic effects should be included in the modeling. The trajectory generation methods considered for an EOAM place an even higher demand on real-time processing capabilities, due to higher ARV speeds, and large-magnitude constraints for lateral acceleration, compared to non-emergency maneuvers (Paden, Čáp, Yong, Yershov, & Frazzoli, 2016), (Claussmann, Revilloud, Gruyer, & Glaser, 2019).

## Kinodynamic Considerations in Trajectory Generation

In addition to the emphasis on real-time processing and controller outputs, dynamic ARV EOAM trajectories need to consider the non-holonomic (Whittaker, 1904), (Laumond, Sekhavat, & Lamiraux, 1998) and kinodynamic (Canny, Donald, Reif, & Xavier, 1988) natures of an ARV completing such maneuvers. Within the Kinodynamic Considerations section of this manuscript (above), it was discussed that non-holonomic considerations limit the available kinematic motions of the ARV, and due to the number of system DOFs exceeding the number of controllable DOFs, such trajectory must be continuously differentiable with time. These non-holonomic and kinematic capabilities of the ARV are affected by its steering range constraints, steering configuration, tractive force configuration (FWD, RWD or AWD), and overall dimensions. Likewise, the dynamic artifacts of the ARV's kinodynamic nature, relate to steering and steering rate, acceleration, and braking, while considering tire/road friction (Katrakazas, Quddus, Chen, & Deka, 2015), (Paden, Čáp, Yong, Yershov, & Frazzoli, 2016). Additionally, the dynamic considerations in trajectory generation and planning relate to passenger comfort, directly tied to longitudinal and lateral acceleration, as well as jerk[51] (Chee & Tomizuka, 1994). It is also noted that ARV must both be able to adhere to a kinematically and dynamically feasible trajectory with a coupled speed profile, to effectively execute an EOAM (Shiller & Sundar, 1998).

## Common Methods for Use in Time-Sensitive Trajectory Generation

To adhere to the non-holonomic (Whittaker, 1904), (Laumond, Sekhavat, & Lamiraux, 1998) and kinodynamic constraints of ARV motion during an EOAM, the most commonly considered trajectory generation approaches are: Parametric curves, Geometric curves, and Function Optimization Methodologies (González, Pérez, Milanés, & Nashashibi, 2015), (Katrakazas, Quddus, Chen, & Deka, 2015), (Sharma, Sahoo, & Puhan, 2019), (Claussmann, Revilloud, Gruyer, & Glaser, 2019), (Zhu & Aksun-Guvenc, Trajectory Planning of Autonomous Vehicles Based on Parameterized Control Optimization in Dynamic on-Road Environments, 2019), (Zhu S. , Gelbal, Aksun-Guvenc, & Guvenc, 2019). These available trajectory generation methods are notable for their relative simplicity and smoothness[52] (Barsky & DeRose, 1984) and availability of numerical solutions. Ideally, these calculation traits will also translate to quickness in computation, to achieve real-time outputs.

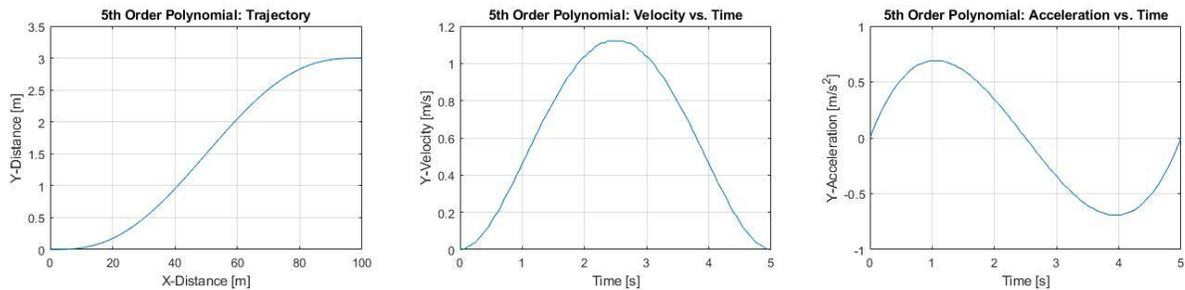

*Figure 52: Example fifth order polynomial trajectory (with constant longitudinal speed profile), which meets relevant kinodynamic requirements for an ARV EOAM*

The state-of-the art Parametric (interpolating) Curve generation techniques often include methods such as Polynomials (Figure 52) (Nelson, 1989), (You, et al., 2015), (Chen & Huang, 2018), (Wang, Cui, & Yu, 2019) (Ding Y. , Zhuang, Qian, & Zhong, 2019), (Mehmood, Liaquat, Bhatti, & Rasool, 2019), (Li, et al., 2019), Splines (Shiller & Sundar, 1998), (Elbanhawi, Simic, & Jazar, 2015), (Li, Sun, Cao, He, & Zhu, 2016), Bézier curves (Bae, Moon, Park, Kim, & Kim, 2013), (González, Pérez, Lattarulo, Milanés, & Nashashibi, 2014), where the parameter most often utilized is time. Geometric curves utilize arclength as to describe their geometry, and these trajectory generation methods include Clothoids[53] (Funke, et al., 2012), (Funke & Gerdes, 2016), (Lundberg, 2017), (Silva & Grassi, 2018), Spirals (Cook, 1979), and Splines (Piazzi & Bianco, 2000), (Piazzi A. , Bianco, Bertozzi, Fascioli, & Broggi, 2002), (Levien, 2009), (Yang K. , 2013),  (Gu & Dolan, 2014), (Zhu, Gelbal, & Aksun-Guvenc, 2018). The most common contemporary Function Optimization Methodologies used in trajectory generation for ARVs are various types of Optimal Control (Chee & Tomizuka, 1994), (Bevan, Gollee, & O'reilly, 2010), (Subosits & Gerdes, 2019), and Model Predictive Control (MPC)[54] (Keviczky, Falcone, Borrelli, Asgari, & Hrovat, 2006), (Falcone, Borrelli, Tseng, H. E., & Hrovat, 2008), (Choi, Kang, & Lee, 2012). With the MPC trajectory generation methods even though the prediction horizon and control horizon consist of several discrete time steps, the entire optimization to goal is redone at every time step until the maneuver is complete (Katrakazas, Quddus, Chen, & Deka, 2015). Additionally, with MPC, all control constraints are considered simultaneously. While this characteristic of MPC is useful for following accuracy, and EOAMs with dynamic objects, it can also be computationally heavy (Sharma, Sahoo, & Puhan, 2019). Each of these methods listed are well-suited for highly dynamic ARV maneuvers such as EOAMs, but also have important tradeoffs to consider in the ARV EOAM logic design (Amer, Zamzuri, Hudha, & Kadir, 2017), (González, Pérez, Milanés, & Nashashibi, 2015), (Paden, Čáp, Yong, Yershov, & Frazzoli, 2016).

---

[51] In dynamics, jerk is defined as the time rate of change of acceleration, of a point (Gillespie, 1992)

[52] Smooth in this sense means that it lacks discontinuities over the interval on which the ARV will execute the EOAM

[53] Also known as Euler or Cornu spirals (Vázquez-Méndez & Casal, 2016)

### *Parametric and Geometric Trajectory Generation Details*

When considering the parametric and geometric trajectory generation methods for an ARV, it is important to note the continuity (Levien, 2009) of these types of trajectories. Parametric continuity is associated with the trajectory and its parameterization, such that the parametrization and its $n$ derivatives agree where the curve segments join, in which that trajectory has $C^n$ continuity (Barsky & DeRose, 1984). Parametric continuity of the trajectory meeting point tangents, speed or first derivative is $C^1$ continuity, whereas $C^2$ continuity denotes equivalent second derivatives or accelerations of the trajectory joining points, $C^3$ denotes equivalent third derivatives or jerk. In other words, time derivatives of the trajectory equations are considered when defining parametric continuity.

Geometric continuity includes a parameterization coupled with constraints that control the shape of the trajectory, and particularly maintaining a continuous slope ($G^1$ continuity: (Dubins, 1957), (Reeds & Shepp, 1990), (Fraichard & Scheuer, 2004), curvature ($G^2$ continuity: (Piazzi & Bianco, 2000), (Piazzi A. , Bianco, Bertozzi, Fascioli, & Broggi, 2002), (Zhu, Gelbal, & Aksun-Guvenc, 2018)), curvature rate which is associated with fairness[55] ($G^3$ continuity: (Oliveira, Lima, Cirillo, Mårtensson, & Wahlberg, 2018), (Banzhaf, Berinpanathan, Nienhüser, & Zöllner, 2018)[56] ), and curvature acceleration ($G^4$ continuity), (Banzhaf, Berinpanathan, Nienhüser, & Zöllner, 2018) (Levien, 2009). When defining $G^n$ continuity, the $n$ denotes the highest order of the derivative for the shape parameter used to define the trajectory (Barsky & DeRose, 1984). When considering ARVs, arclength is commonly used as the shape parameter to define geometrically continuous trajectories, especially when the parametric equation considered is heading angle with respect to arclength (Banzhaf, Berinpanathan, Nienhüser, & Zöllner, 2018).

With both Parametric and Geometric trajectory generation methods, the continuity and associated number can be used to describe the robustness of the methodology. However, as the continuity number increases for Parametric or Geometric trajectory methods, so does the computational complexity, and severity of constraints. This can result in longer trajectory generation computation times, but special methods can be constructed to reduce the required computation times (Banzhaf, Berinpanathan, Nienhüser, & Zöllner, 2018), (Oliveira, Lima, Cirillo, Mårtensson, & Wahlberg, 2018). It should also be noted that of the above examples, (Shiller & Sundar, 1998) utilized a Parametric or Geometric method for the initial reference profile of the ARV EOAM, and then utilized Function Optimization Methodologies to determine the final states and speed profiles during the maneuver (Figure 53).

---

[54] The trajectory generation within the MPC method will be discussed in the Trajectory-Tracking Control Strategies, since the trajectory generation and the control strategy are closely coupled, with MPC.

[55] Fairness is synonymous with the term smoothness of curve, which equates to the variation of curvature, and fairness is not equivalent to continuity (Levien, 2009)

[56] It should be noted that several Geometric-based methodologies for trajectory generation of a nonholonomic vehicle, such as those by (Oliveira, Lima, Cirillo, Mårtensson, & Wahlberg, 2018), (Banzhaf, Berinpanathan, Nienhüser, & Zöllner, 2018), (Silva & Grassi, 2018) assume vehicle movement at speeds where a kinematic vehicle model may be utilized where this is not the case for EOAMs

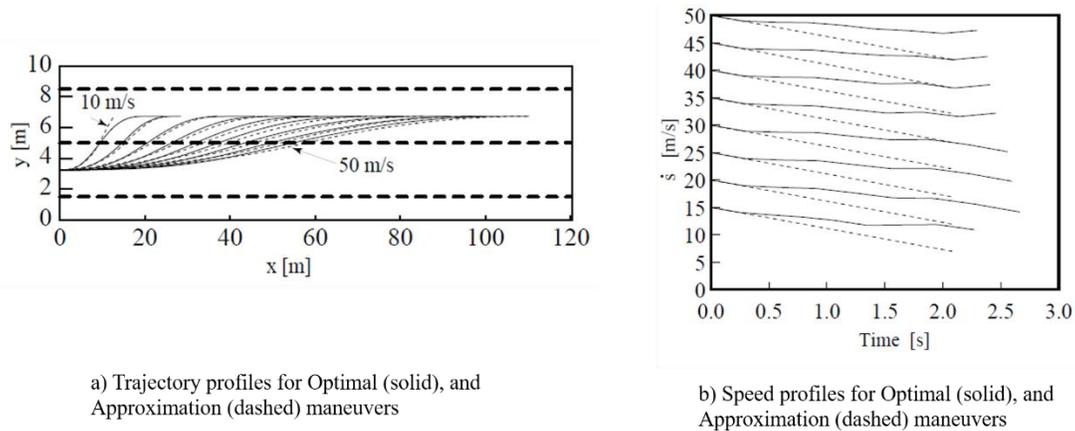

a) Trajectory profiles for Optimal (solid), and
Approximation (dashed) maneuvers

b) Speed profiles for Optimal (solid), and
Approximation (dashed) maneuvers

*Figure 53: Comparison for Optimal (Bicycle Vehicle Model) vs. Approximated (Point-Mass Vehicle Model) a) Trajectory and b) Speed profiles for the maneuver (Shiller & Sundar, 1998)*

## Human Driver Performance-Based Trajectory Generation

One naturalistic approach for trajectory generation is to generate trajectories based on human driving data (Gu & Dolan, 2014), (Peng, et al., 2020). In each of these approaches the goal was to mimic human driving behaviors as references for driving inputs for an ARV. The method (Gu & Dolan, 2014) took was to capture actual on-road data from human drivers in urban and highway scenarios for reference data, then interpolate, smooth, and reduce the curvature of the steering data using a Geometric-based approach. After that they used that reference steering geometry to create an associated speed profile.

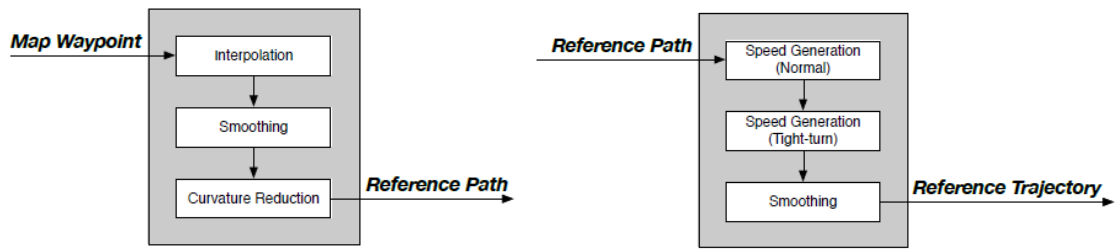

*Figure 54: human-based trajectory generation by (Gu & Dolan, 2014)*

Alternatively, human driving data was captured from a driving simulator, with participants who were tasked with performing an obstacle avoidance maneuver under specific conditions on both a straight and curved road. The human driving inputs for the EOAM were then used by (Peng, et al., 2020) to form an applied Gaussian fit to the lateral velocity data, to form a parameterized reference trajectory for an ARV (Figure 55).

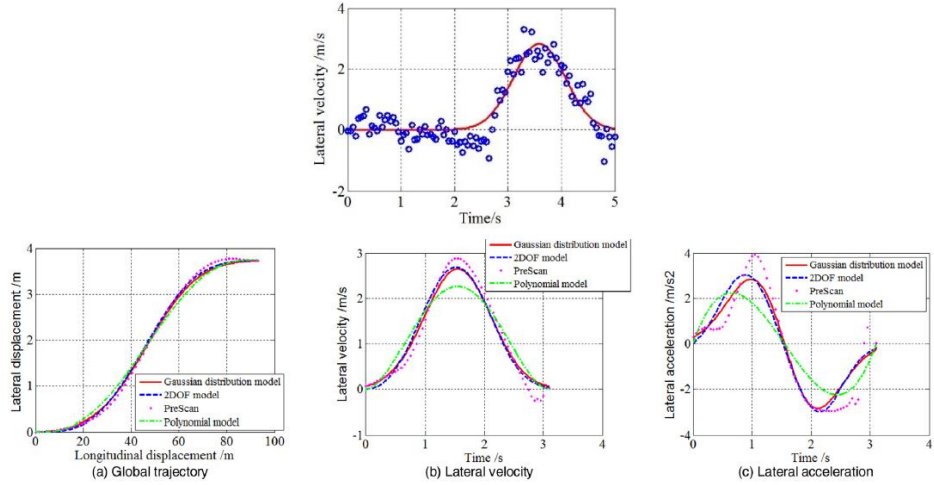

*Figure 55: Gaussian fit of driver EOAM lateral velocity data, and associated trajectory comparisons (Peng, et al., 2020)*

This method was also notable due to inclusion of a special parameter within the Gaussian-based trajectory formulation that could be adjusted to account for various nominal level of vehicle capabilities (i.e., sports car versus truck), and even changing tire-road friction levels (Figure 56). It should be noted that for these experiments, constant longitudinal velocities were chosen for the duration of the EOAM.

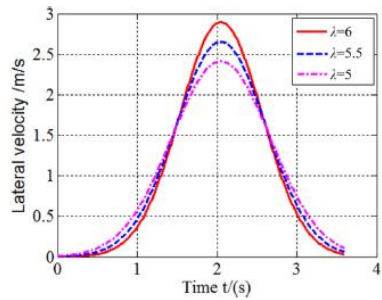

Fig. 3. Vehicle lateral velocity variation in the time domain with $\lambda$.

*Figure 56: Illustration of lateral velocity basis curve adjustment by parameter $\lambda$ (Peng, et al., 2020)*

With the potential ARV EOAM trajectory generation methods noted above, the shape of the trajectory used to avoid the obstacle – with considerations of continuity and smoothness – as well as computational complexity (and associated computation time), should be considered in choosing a preferred method for a given ARV EOAM framework.

## Trajectory-Tracking Control Strategies

The control methodologies used to provide ARV actuator inputs (steering, throttle, brakes) during an EOAM, are based on the previously generated trajectory, and current vehicle states. Specifically, there must be logic implemented which controls the ARV's lateral control (steering), and longitudinal control (speed/acceleration from traction power source or braking). These controls can be feedforward and/or feedback (Talvala & Gerdes, 2008), and the latter involves the goal of minimizing the error between the reference inputs and actual ARV output (Jo K. , Kim, Kim, Jang, & Sunwoo, 2015)  (Figure 57).

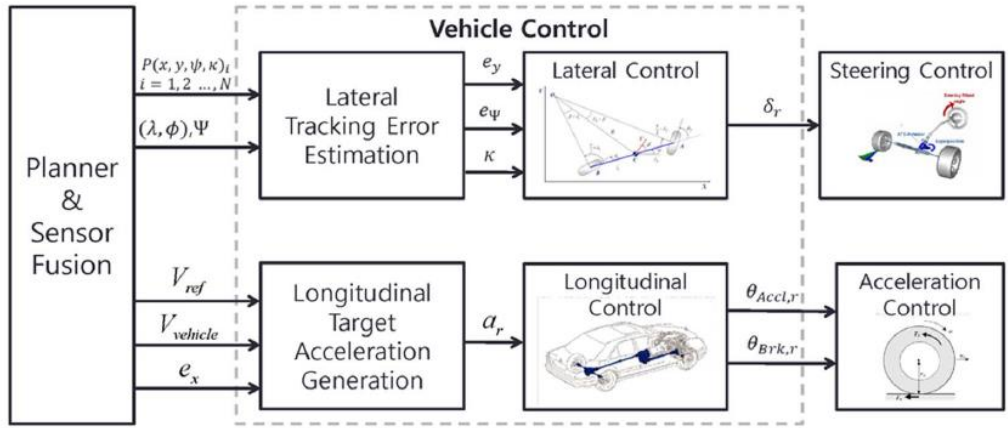

*Figure 57: Basic layout for lateral and longitudinal control of an ARV (Jo K. , Kim, Kim, Jang, & Sunwoo, 2015)*

There are two prominent reviews of path and trajectory-tracking controller methodologies, by (Paden, Čáp, Yong, Yershov, & Frazzoli, 2016) and (Amer, Zamzuri, Hudha, & Kadir, 2017), which have summaries that can be seen in (Figure 58) and (Figure 59), respectively.

| Controller | | Model | Stability | Time Complexity | Comments/Assumptions |
|---|---|---|---|---|---|
| Pure Pursuit | (V-A1) | Kinematic | LES* to ref. path | $O(n)^\star$ | No path curvature |
| Rear wheel based feedback | (V-A2) | Kinematic | LES* to ref. path | $O(n)^\star$ | $C^2(\mathbb{R}^n)$ ref. paths |
| Front wheel based feedback | (V-A3) | Kinematic | LES* to ref. path | $O(n)^\star$ | $C^1(\mathbb{R}^n)$ ref. paths; Forward driving only |
| Feedback linearization | (V-B2) | Steering rate controlled kinematic | LES* to ref. traj. | $O(1)$ | $C^1(\mathbb{R}^n)$ ref. traj.; Forward driving only |
| Control Lyapunov design | (V-B1) | Kinematic | LES* to ref. traj. | $O(1)$ | Stable for constant path curvature and velocity |
| Linear MPC | (V-C) | $C^1(\mathbb{R}^n \times \mathbb{R}^m)$ model‡ | LES* to ref. or path | $O\left(\sqrt{N}\ln\left(\frac{N}{\varepsilon}\right)\right)^\dagger$ | Stability depends on horizon length |
| Nonlinear MPC | (V-C) | $C^1(\mathbb{R}^n \times \mathbb{R}^m)$ model‡ | Not guaranteed | $O(\frac{1}{\varepsilon})^\ddagger$ | Works well in practice |

Table II: Overview of controllers discussed within this section. **Legend:** ⋆: local exponential stability (LES); ∗: assuming (V.1) is evaluated by a linear search over an $n$-point discretization of the path or trajectory; †: assuming the use of an interior-point method to solve (V.28) with a time horizon of $n$ and solution accuracy of $\varepsilon$; ‡: based on asymptotic convergence rate to local minimum of (V.25) using steepest descent. Not guaranteed to return solution or find global minimum.; ‡: vector field over the state space $\mathbb{R}^n$ defined by each input in $\mathbb{R}^m$ is a continuously differentiable function so that the gradient of the cost or linearization about the reference is defined.

*Figure 58: A summary of controller types for ARV path and trajectory tracking (Paden, Čáp, Yong, Yershov, & Frazzoli, 2016)*

| Controller Types | Strength | Weakness | Comments |
|---|---|---|---|
| Geometric & Kinematic | Relatively the simplest type of controller due to least complex state variables | Does not regard the dynamics of the vehicle in controller | Most suitable in applications where dynamics of the vehicle can be ignored |
| | | Controller parameters can be over-tuned and path dependent | |
| Dynamic | Dynamic effect of the vehicle is included in control law | Attainment of vehicle's dynamic states (e.g. Forces, torques) may not be straightforward in experiment. | Most suitable in applications where dynamics of the vehicle is crucial to control and the computational power on-board is sufficient to obtain and process dynamic states |
| Optimal | Certain optimal controller (e.g. LQR) carry out offline gain optimisation →simple online controller | Some optimal control may require online optimisation (e.g. Sharp et al. [80]) and can be computationally demanding | Suitable for robust applications |
| | | Development of the controller is based on linear assumptions. May limit controller's ability | |
| Adaptive | Adapt to various operating conditions making the controller robust to changes | Robustness may be catered only for a specific condition | |
| | If intelligence algorithms (e.g. NN or fuzzy) is included, will simplify controller and need less computational effort | Some adaptive algorithms will be complex and hence, increasing the computational cost of the overall approach | |
| Model Based | Consider the overall vehicle model in determining control signals | Involve online optimisation problem within the controller. Heuristics method will require a computational resources | |
| | Adapt to changes in vehicle parameters making the controller more robust | | |
| Classical | Established method in control field especially good in nonlinear system control. | May need complex derivations and selections (e.g. SMC and H infinity) | |
| | Solution to common problems can be found in standard literature | | |

*Figure 59: A summary of controller types for ARV path and trajectory tracking (Amer, Zamzuri, Hudha, & Kadir, 2017)*

With the work by (Paden, Čáp, Yong, Yershov, & Frazzoli, 2016), the V-AX category denotes controller strategies for path stabilization, V-BX is for trajectory stabilization, and V-C is for trajectory stabilization for complex vehicle models. With respect to EOAMs for ARVs the V-BX and V-C controller categories are most useful due to the dynamic nature of the necessary maneuver and relevant vehicle models. The V-BX and V-C controller categories include dynamic properties of the utilized vehicle models, in the controllers themselves. Likewise, in (Amer, Zamzuri, Hudha, & Kadir, 2017), the Dynamic, Optimal, Adaptive, and Model-Based controller categories include the dynamic properties of the vehicles within the controllers. These controllers also include methods of stabilization such as feedback linearization (d'Andréa-Novel, Campion, & Bastin, 1995), (De Luca, Oriolo, & Samson, 1998), (Peng, et al., 2020), and Lyapunov-based stabilization[57]: (Benton & Smith, 2005), (Talvala & Gerdes, 2008), (Talvala, Kritayakirana, & Gerdes, 2011). Additionally, dynamic control methods have been known to adopt parameter space robust control to help achieve the desired dynamic performance with a wide range of parameter uncertainty: (Ackermann, Guldner, Sienel, Steinhauser, & Utkin, 1995), (Zhu, Gelbal, & Aksun-Guvenc, 2018), (Zhu S. , Gelbal, Aksun-Guvenc, & Guvenc, 2019), (Guvenc, Guvenc, Demirel, & Emirler, 2017).

Certain Dynamic controller methods aim to address nonlinearities that exist directly in the vehicle model, or indirectly in the tire model such as the previously mentioned feedback linearization, in addition to methods such as Adaptive Control (Figure 60) and Gain Scheduling. Adaptive control (Amer, et al., 2018), and gain scheduling (Wang, Montanaro, Fallah, Sorniotti, & Lenzo, 2018), (Zhu S. , Gelbal, Aksun-Guvenc, & Guvenc, 2019) methods both linearize the system dynamics throughout the trajectory-tracking control, however, gain scheduling does this at user-defined and relevant operating points (Paden, Čáp, Yong, Yershov, & Frazzoli, 2016), whereas adaptive control completes this at each time step (Amer, Zamzuri, Hudha, & Kadir, 2017). The advantage of gain scheduling is that the computation time is lower than adaptive control (generally), but more computational memory is needed to store information regarding the ARV system linearization at each operating point (Amer, Zamzuri, Hudha, & Kadir, 2017). Though adaptive control can have higher computation time than gain scheduling, it requires less memory storage, due to the online linearization of the system.

---

[57] This is stabilization based on the canonical Lyapunov function.

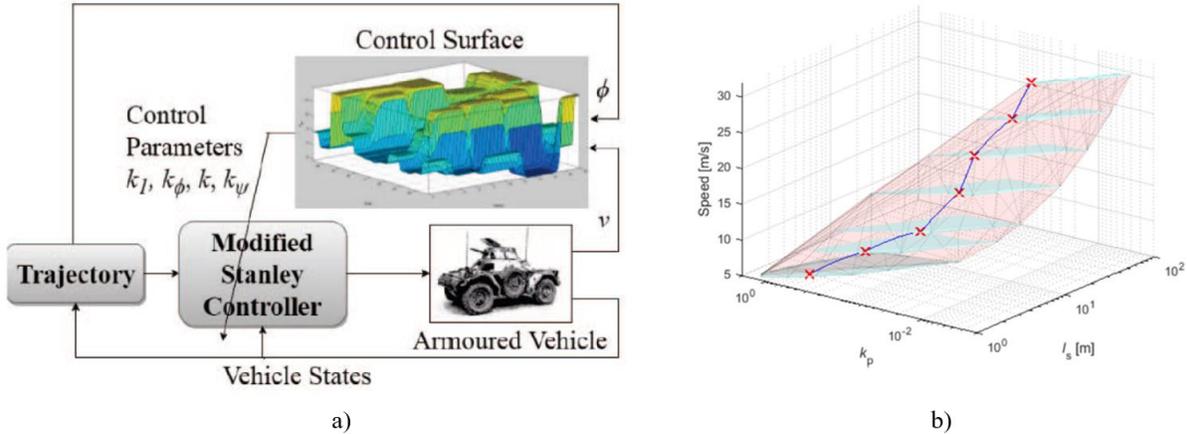

a)                                                                  b)

*Figure 60: (a) Examples of an Adaptive Control system (Amer, et al., 2018), and (b) Gain Scheduling control law output parameters $l_s$, and $k_p$) (Zhu S. , Gelbal, Aksun-Guvenc, & Guvenc, 2019)*

Nonlinear controller methods that have been previously mentioned that are applicable for ARVs performing EOAMs can be various types of Function Optimization methods, MPC or nonlinear MPC (NMPC)[58]. As mentioned in the Common Methods for Use in Time-Sensitive Trajectory Generation section, these nonlinear control methods can be superior in terms of trajectory-tracking accuracy for applications like EOAMs, but can suffer in terms of computational load (Paden, Čáp, Yong, Yershov, & Frazzoli, 2016), (Amer, Zamzuri, Hudha, & Kadir, 2017), (Sharma, Sahoo, & Puhan, 2019). Still there are various applications of nonlinear control that are shown to be useful for ARV EOAM applications, given reasonable computational power available, to achieve real-time outputs. Some of these applications can be categorized in the general Optimal Control: (Bevan, Gollee, & O'reilly, 2010), (Ziegler et al., 2014) , (Fahmy, Abd El Ghany, & Baumann, 2018),  (Fors, Olofsson, & Nielsen, 2018), (Subosits & Gerdes, 2019), the specialized case of Quadratic Programming: (Chee & Tomizuka, 1994), (Wang, Montanaro, Fallah, Sorniotti, & Lenzo, 2018) MPC: (Keviczky, Falcone, Borrelli, Asgari, & Hrovat, 2006), (Falcone, Borrelli, Asgari, Tseng, & Hrovat, 2007), (Falcone, Borrelli, Tseng, H. E., & Hrovat, 2008), and NMPC: (Gao, Lin, Borrelli, Tseng, & Hrovat, 2010), (Choi, Kang, & Lee, 2012), (Attia, Orjuela, & Basset, 2012).

While Nonlinear MPC is certainly an option for trajectory-tracking control it is also the most computationally expensive one (Paden, Čáp, Yong, Yershov, & Frazzoli, 2016), (Claussmann, Revilloud, Gruyer, & Glaser, 2019). However, with the use of gain scheduling, adaptive control, or other linearization techniques, Nonlinear MPC can be reduced to Linear MPC which can have convex solutions, under certain conditions. In a similar manner, certain classical control methods such as proportional, integral, derivative (PID): (Tseng, et al., 2005), (Hoffmann, Tomlin, Montemerlo, & Thrun, 2007),  (Kritayakirana & Gerdes, 2012) (Mehmood, Liaquat, Bhatti, & Rasool, 2019), and neural network (Chen & Huang, 2018) (Ding Y. , Zhuang, Qian, & Zhong, 2019), or Sliding Mode Control: (Chee & Tomizuka, 1994), (Ackermann, Guldner, Sienel, Steinhauser, & Utkin, 1995), (Ackermann J. , 1980), (Wang, Steiber,

---

[58] NMPC is differentiated from MPC in that NMPC includes nonlinear optimization routines whereas MPC includes linear optimization routines

& Surampudi, 2008), (Wang, Cui, & Yu, 2019), can be used when the feedback linearization or gain scheduling methods are also included.

# Conclusion

As autonomous road vehicles (ARVs) advance, they must transcend general road and traffic conditions, to some of the more extreme conditions that human drivers can also face. The ability to successfully navigate circumstances that require emergency obstacle avoidance maneuvers (EOAMs) while autonomously operating, is a necessary capability to instill full confidence in ARVs. The literature regarding the four general ARV systems – sensing and perception, decision-making, path/trajectory generation, and path/trajectory-tracking control – were reviewed in this paper, with special emphasis on the latter three, and especially in the context of an EOAM at highway speeds while placing special emphasis on the required combination and successful function of all four systems, when considering EOAMs at highway speeds.

This combination of all four systems and their cooperative functions is important to consider because an actual ARV cannot execute an EOAM without the following: sensors that obtain useful environmental data, onboard processors that perceive those sensor data as relevant outputs for the ARVs subsystems, properly generated emergency trajectories given the environment and vehicle state, and control systems that can manage the ARV actuators to follow the low-level reference inputs.

With emphasis on harmonizing the four general ARV system outputs to work together for a successful EOAM, this review paper offered a novel perspective on successful EOAM performance, with special considerations for EOAMs at higher speeds or lower tire-road surface friction conditions.

# Recommendations for Future Work

While this review offered a novel perspective on how an EOAM framework for ARVs could fit within the general ARV system layout, it stopped short of specifying unique details of an EOAM framework that considers relevant interactions with the other general systems of an ARV at highway speeds. Some of these details might consider a comparison of trajectory-generation and trajectory-tracking control methods to understand which ones provide a good compromise of real-time performance, robustness to parameter changes, handling of multiple object-road encroachment modes, and differences between operation at highway speeds versus lower speeds with lower tire-road surface friction.

This review also mentioned aspects of EOAM subsystem performance that would be necessary to complete an EOAM, especially at higher speeds, However, it did not specify how appropriate testing judgment of an ARV's EOAM performance could be manifested in a regular and specific protocol. This protocol would place emphasis on the successfully collaboration of all four general ARV systems (Sensing and Perception, Motion Planning/Decision-Making, Path Generation, and Control), throughout the duration of an EOAM. Such a protocol has been mentioned briefly before, but not yet developed in a significant manifestation (Lowe, Zhu, Aksun-Guvenc, & Guvenc, 2019).

These are two primary areas of potential future work, which could solidify the overall Level 3-5 passenger ARV performance, by assuring an ARV would be prepared for some edge conditions that a human driver might face, such as an EOAM.

One area that this review did not address but is relevant to ARV EOAMs is ethics in EOAM algorithm creation. The topic of ethics in ARV driving and algorithm creation is a topic that has been broached previously (Goodall, 2014), (Gerdes & Thornton, 2015), (Hevelke & Nida-Rümelin, 2016), (Lin, 2016), (Bonnefon, Shariff, & Rahwan, 2016), (Fleetwood, 2017), (Claussmann, Revilloud, Gruyer, & Glaser, 2019). This is important when considering the case when an impact of some nature has been calculated as imminent. When a collision is imminent with an ARV, what or whom will the algorithm decide is most important to limit damage? This and other related ethical concerns in implementation of an EOAM for an ARV must be explored in more detail.